\documentclass[10pt,journal,compsoc]{IEEEtran}
\ifCLASSOPTIONcompsoc
  \usepackage[nocompress]{cite}
\else
  \usepackage{cite}
\fi
\ifCLASSINFOpdf
\else
\fi
\ifCLASSOPTIONcompsoc
 \usepackage[caption=false,font=footnotesize,labelfont=sf,textfont=sf]{subfig}
\else
 \usepackage[caption=false,font=footnotesize]{subfig}
\fi
\usepackage{times}
\usepackage{epsfig}
\usepackage{graphicx}
\usepackage{amsmath}
\usepackage{amssymb}

\usepackage{tablefootnote}
\usepackage{multirow}
\usepackage{xcolor} 
\usepackage{verbatim}
\usepackage{url}
\usepackage{bm}
\usepackage{caption}
\usepackage{adjustbox}
\usepackage{balance}
\usepackage{mathrsfs}
\usepackage{pifont}
\usepackage{diagbox}
\usepackage{float}
\usepackage{dblfloatfix}
\newcommand{\cmark}{\ding{51}}%
\newcommand{\xmark}{\ding{55}}%
\newcommand{\myparagraph}[1]{\vspace{0.8em}\noindent\textbf{#1}}
\newcommand{\etal}{\textit{et al.}}

\makeatletter
\long\def\@IEEEtitleabstractindextextbox#1{\parbox{0.922\textwidth}{#1}}
\makeatother

\begin{document}
%
\title{Cascaded Refinement Network for Point Cloud Completion with Self-supervision}
%
%
%
%

\author{Xiaogang Wang,
        Marcelo H Ang Jr,~\IEEEmembership{Senior Member,~IEEE,} 
        and~Gim Hee Lee,~\IEEEmembership{Member,~IEEE}
\IEEEcompsocitemizethanks{
\IEEEcompsocthanksitem X. Wang is with the Department
of Mechanical Engineering, National University of Singapore, Singapore, 117575. E-mail: xiaogangw@u.nus.edu\protect\\
\IEEEcompsocthanksitem M. H. Ang, Jr. is with the Advanced Robotics Centre, Department of Mechanical Engineering, c, 117608. E-mail: mpeangh@nus.edu.sg \protect\\
\IEEEcompsocthanksitem  G. H. Lee is with the Department of Computer Science, National University of Singapore, Computing 1, 13 Computing Drive, Singapore.
E-mail: gimhee.lee@nus.edu.sg 
}
\thanks{Manuscript received August, 2020}}

\IEEEtitleabstractindextext{%
\begin{abstract}
Point clouds are often sparse and incomplete, which imposes difficulties for real-world applications. Existing shape completion methods tend to generate rough shapes without fine-grained details. Considering this, we introduce a two-branch network for shape completion. The first branch is a cascaded shape completion sub-network to synthesize complete objects, where we propose to use the partial input together with the coarse output to preserve the object details during the dense point reconstruction. The second branch is an auto-encoder to reconstruct the original partial input. The two branches share a same feature extractor to learn an accurate global feature for shape completion. Furthermore, we propose two strategies to enable the training of our network when ground truth data are not available. This is to mitigate the dependence of existing approaches on large amounts of ground truth training data that are often difficult to obtain in real-world applications. Additionally, our proposed strategies are also able to improve the reconstruction quality for fully supervised learning. We verify our approach in self-supervised, semi-supervised and fully supervised settings with superior performances. Quantitative and qualitative results on different datasets demonstrate that our method achieves more realistic outputs than state-of-the-art approaches on the point cloud completion task.
\end{abstract}

}

\maketitle

\IEEEdisplaynontitleabstractindextext

\ifCLASSOPTIONpeerreview
\begin{center} \bfseries EDICS Category: 3-BBND \end{center}
\fi
%
\IEEEpeerreviewmaketitle

\IEEEraisesectionheading{\section{Introduction}\label{sec:introduction}}

%
%
%
%
\IEEEPARstart{3D} point cloud 
is increasingly used as a flexible representation in computer vision due to the 
availability of various easily accessible commercially off-the-shelf depth scanning devices.
Numerous works~\cite{qi2017pointnet,qi2017pointnet++,li2018so,dgcnn} have been proposed for point cloud analysis by directly extracting point-wise features from the point coordinates. Although those approaches have achieved impressive 3D object classification or semantic segmentation results, they are designed for complete and dense point sets, and show inferior performances when applied to incomplete and sparse 3D shapes, such as the raw data acquired by Lidars in self-driving cars and autonomous drones. The shape completion task whose objective is to reconstruct complete and dense point sets given incomplete and sparse point clouds is therefore crucial for improving the perception performances in object classification~\cite{qi2017pointnet,qi2017pointnet++,dgcnn,achituve2021self}, detection~\cite{lang2019pointpillars,liang2018deep}, tracking~\cite{giancola2019leveraging}, scene understanding~\cite{hou2019sis,dai2018scancomplete} and augmented reality~\cite{boud1999virtual,webster1996augmented}.

\begin{figure}
  \includegraphics[width=\linewidth]{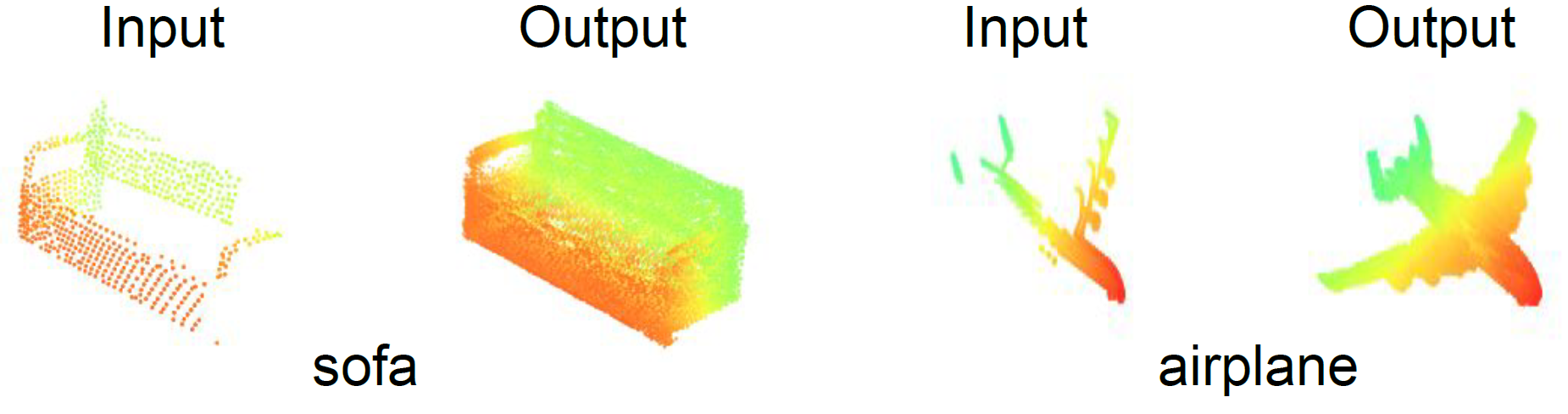}
  \caption{We generate dense and complete objects given sparse and incomplete scans. The resolutions of input and output are 2,048 and 16,384, respectively.}
  \vspace{-3mm}
  \label{task_picture}
\end{figure}

Existing methods~\cite{yuan2018pcn,topnet2019,brock2016generative,dai2017shape,litany2018deformable,sinha2017surfnet} have shown promising results on shape completion for different inputs, such as voxel grids, meshes and point clouds. 
Voxel representation is a direct generalization of pixels to the 3D case.
Although it is straightforward to adapt 2D convolutions to 3D~\cite{wu2016learning,smith2017improved}, the reconstruction process suffers from memory inefficiency, and thus it is difficult to obtain high-resolution results. 
Data-driven methods on mesh representations~\cite{wang2018pixel2mesh,groueix2018papier,wang20193dn} show superiority on generating complicated surfaces with less memory. However, it is difficult to change the topology during training since they are limited to the fixed vertex connection patterns.
In contrast, it is easy to add new points for point clouds, and several studies have shown promising results.
The pioneering work PCN~\cite{yuan2018pcn} proposes an encoder-decoder architecture for point cloud completion on synthetic and real-world datasets. 
A following work TopNet~\cite{topnet2019} proposes a hierarchical decoder with a rooted tree structure to generate dense and complete objects. Although PCN and TopNet have achieved impressive performances on different resolutions, they are unable to generate detailed geometric structures of 3D objects.
To alleviate this problem, several approaches~\cite{mescheder2019occupancy,park2019deepsdf,michalkiewicz2019deep} propose to learn 3D structures in a function space, where they predict a binary classification score for each cell in a volumetric space and then reconstruct surface boundaries by marching cube algorithm~\cite{cubes1987high}.
Although these methods have achieved impressive results, they are all supervised learning approaches that require large amounts of difficult to obtain ground truth training data. 
In this work, we first propose a two-branch point completion network to generate objects with fine-grained details. 
We then introduce two strategies to train the network with a self-supervised training scheme.
We further show that our self-supervised strategies are able to improve fully supervised learning with different backbones.
Some results of our approach are shown in Figure~\ref{task_picture}.

Our two-branch completion network includes a point completion branch and a partial reconstruction branch.
We design the point completion branch in a coarse-to-fine manner to generate fine-grained object details. We propose to integrate the partial input with the coarse outputs in the second stage to preserve local details from the incomplete points. However, simple concatenation between the partial inputs and coarse outputs gives rise to unevenly distributed points (shown in the 3rd column of Figure~\ref{iterative_results}). To alleviate this problem, we design an iterative refinement decoder to gradually refine the point coordinates when generating the missing parts. Specifically, we split the generation process into several iterations instead of generating the dense and complete point clouds in a single forward step. This iterative refinement strategy makes it easier to train neural networks and reduces the parameters since the parameters are shared among different iterations.
The partial reconstruction branch is an auto-encoder sub-network.  
By sharing weights of the feature extractor of the two branches, we are able to preserve critical point features for the shape completion sub-network.

Given that large amounts of ground truth is not always available 
for training, we propose two self-supervised training strategies: 1) partial resampling; 2) point cloud MixUp inspired by~\cite{achituve2021self}. 
We sample a more incomplete point set from the original partial input in the resampling strategy. This strategy aims to mimic the standard procedure of shape completion by correlating two different levels of partialness from the same object.
In the MixUp strategy, we generate more training pairs by taking the convex combination of two different object instances. 
The proposed two strategies enable our network to be trained not only with a self-supervised training scheme, but also in a fully supervised fashion. 
This is verified by the improved performances of our network in the supervised, semi-supervised and self-supervised settings. 

We conduct extensive experiments on different datasets to verify the effectiveness of our proposed approach. Our key contributions are summarized as follows:
\begin{itemize}
    \item We design a two-branch shape completion network that is able to preserve object details from partial points and concurrently generate missing parts with high-quality.
    
    \item We propose a cascaded refinement strategy to gradually refine point locations locally and globally and use it together with the coarse-to-fine pipeline.
    
    \item We propose two self-training strategies to improve the reconstruction performances in both supervised and self-supervised settings.
    
    \item Our proposed self-supervised strategies can be incorporated into existing completion backbones seamlessly and improve their performances consistently.
    
    \item Experiments on different datasets show that our framework achieves superior results than existing methods on the 3D point cloud completion task.
\end{itemize}

\section{Related works}
We review existing works on point cloud analysis and related works on shape completion in this section.

\subsection{Point Cloud Analysis}
Point cloud learning has attracted increasing attention these years, and a large number of works have been proposed for related tasks like classification, segmentation and tracking. Currently, point cloud analysis methods can be roughly categorized into three directions~\cite{guo2020deep}: point-wise Multi-Layer Perceptrons (MLP)~\cite{qi2017pointnet} related methods, convolution-based methods and graph-based methods.  

\myparagraph{MLP-based Methods.} Point-wise MLP based methods extract each point features independently by several shared MLPs and obtain the global object features via a symmetric function such as average-pooling or max-pooling. The pioneering work PointNet~\cite{qi2017pointnet} extracts the point-wise features by several MLPs and obtains the global embedding by a max-pooling operation. Although PointNet achieves impressive results on object classification and segmentation, they neglect the relationships among different points. Therefore, PointNet++~\cite{qi2017pointnet++} proposes a hierarchical network to model the relationship within a local area by a set abstraction module. 
Several works~\cite{yang2019modeling,zhao2019pointweb,duan2019structural,yan2020pointasnl} are proposed to consider the context of the local neighborhood during the feature extraction inspired by PointNet++.

\myparagraph{Convolution-based Methods.} Comparing with point-wise MLPs, convolution-based methods follow the conventional convolution mechanism in 2D image processing. To generalize typical CNNs to point clouds, PointCNN~\cite{li2018pointcnn} proposes a $\chi$-transformation on the points to transform the unordered points into potentially canonical ordered ones by MLPs. Consequently, point features can be extracted by typical convolutions on the ordered points.
Other methods such as KPConv~\cite{thomas2019kpconv} defines a deformable kernel convolution on several kernel points when applying CNNs on point clouds. PointConv~\cite{wu2019pointconv} treats convolution kernels as nonlinear functions of the local coordinates with weight and density functions. They propose to use MLPs to approximate a weight function and calculates a density value by a kernelized density estimation to re-weight the weights.

\myparagraph{Graph-based Methods.} Graph-based approaches treat points as vertexes and calculate directed edges based on the nearby points. DGCNN~\cite{dgcnn} proposes a dynamic graph edge convolution on point clouds. The neighbors are selected dynamically based on the distances of point features in every layer. Shen \etal~\cite{shen2018mining} proposes a kernel correlation and graph pooling operation to consider the geometric affinities in the neighboring data points. 

Among various point cloud analysis methods, PointNet~\cite{qi2017pointnet} and PointNet++~\cite{qi2017pointnet++} are most widely used in point cloud generation and completion when extracting global features.

\begin{figure*}
\centering
  \includegraphics[width=1\linewidth]{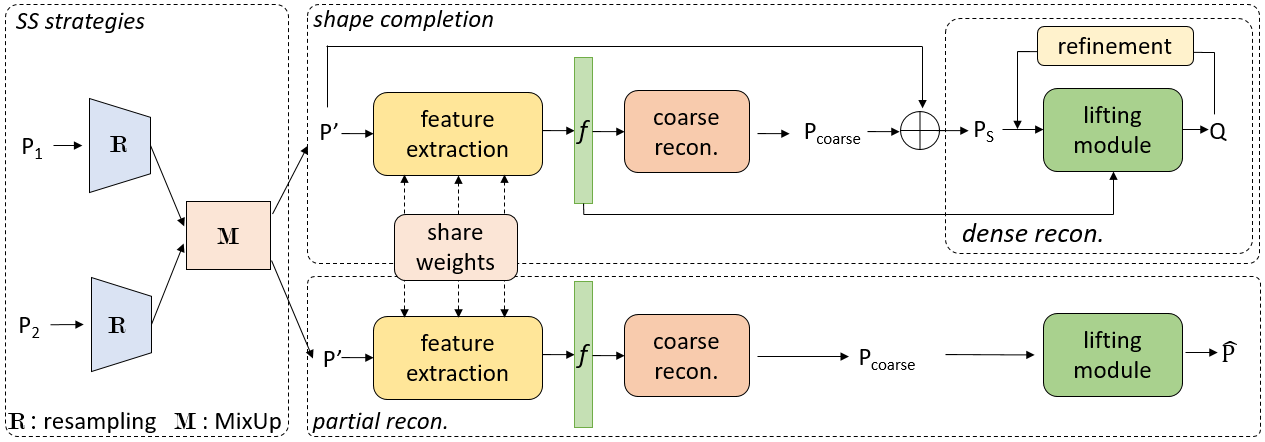}
  \vspace{-5mm}
  \caption{An illustration of our whole pipeline. The entire architecture includes three parts. 
  The left part shows the self-supervised (SS) strategies (Sec.~\ref{ss_strategies}), which includes resampling (Sec.~\ref{resampling}) and MixUp (Sec.~\ref{MixUp}).
  The right top branch is the completion sub-network (Sec.~\ref{Completion_Branch}), which includes three components: feature extraction, coarse reconstruction and dense reconstruction. 
  The dense reconstruction is an iterative refinement sub-network with a lifting module in each step. 
  The right bottom branch is an auto-encoder for the partial reconstruction (Sec.~\ref{Partial_Reconstruction}). It shares the feature extractor with the completion sub-network but removes the skip-connection and refinement step from the shape completion branch.}
  \vspace{-5mm}
  \label{whole_network}
\end{figure*}

\subsection{Point Cloud Completion}
3D shape completion~\cite{dai2017shape,han2017high,stutz2018learning} plays an essential role in robotics and computer vision and has obtained significant development in recent years. Existing methods have shown impressive performances on various representations: voxel grids, meshes and point clouds.
Inspired by 2D CNN operations, earlier works~\cite{dai2017shape,han2017high,le2018pointgrid} focus on the voxel and distance fields formats with 3D convolution. Several approaches~\cite{dai2017shape,han2017high} have proposed a 3D encoder-decoder-based network for shape completion and shown promising performances. However, voxel-based methods consume a large amount of memory and are unable to generate high-resolution outputs. 
To alleviate this problem,~\cite{wang2017cnn,wang2018adaptive} have proposed to use the octree structure to gradually voxelize specific areas to increase the resolution. 
However, recent works gradually discard the voxel format and focus on mesh reconstruction due to the quantization effect of the voxelization operation. 
Nonetheless, existing mesh representations~\cite{vakalopoulou2018atlasnet,wang2018pixel2mesh} are not flexible to arbitrary typologies when deforming a template mesh to a target mesh. 
In comparison to voxels and meshes, it is easier to add new points in point clouds 
during training.
Yuan \etal~\cite{yuan2018pcn} proposes a pioneering work PCN on point cloud completion, which is an encoder-decoder network to reconstruct a dense and complete point set from an incomplete point cloud. They adopt the folding mechanism~\cite{yang2018foldingnet} to generate high-resolution outputs.
Following work TopNet~\cite{topnet2019} proposes a hierarchical tree-structure network to generate point cloud without assuming any specific topology for the input point set. 
However, both PCN and TopNet are unable to synthesize the fine-grained details of 3D objects. 
MSN~\cite{liu2020morphing} proposes a similar conditional method with ours to preserve object details by sampling points from the combined point clouds between a coarse output and the partial input. To preserve more object details, we first sample a subset from the partial input and then concatenate it with the coarse output, which are verified to be able to achieve superior results.

Although recent works~\cite{huang2020pf,zhang2020detail,wen2020point,xie2020grnet,liu2020morphing} on point cloud completion have shown improved performances, they are limited to supervised learning.
SG-NN~\cite{dai2020sg} proposes a self-supervised method on scene completion by removing several frames from a scanned sequence. Subsequently, self-supervision is formulated by correlating two levels of partial scans from the same input sequence. Unlike SG-NN, we remove object regions whose point coordinates are within a distance threshold regarding a randomly sampled center point.

\section{Methodology}
\subsection{Overview}
Given a sparse and incomplete point set $\text{P}=\{p_i\}^\text{N}_{i=1}$ with $\text{N}$ points, the target is to generate a dense and complete point set $\text{Q}=\{q_i\}^{u \times \text{N}}_{i=1}$ with $u \times \text{N}$ points, where $u$ is the upsampling scalar.
More specifically, our objectives are two folds: 
\textbf{(1)} We aim to produce complete and high-resolution 3D objects from corrupted and low-resolution point clouds. The output should preserve local details of the input point cloud $\text{P}$ and concurrently generate the missing parts with detailed geometric structures. 
\textbf{(2)} We aim to train the network with a self-supervised training scheme without utilizing the fully complete ground truth. This is essential for many real-world applications where the ground truth are not available.
In this paper, we refer to point clouds without ground truth as unlabeled data, and point sets with ground truth as labeled data.
We show the overall architecture of our two-branch completion network in Figure~\ref{whole_network}. The right top branch is a coarse-to-fine completion network that generates complete object shapes from partial inputs. The right bottom branch is an auto-encoder to reconstruct the partial inputs. We also show the proposed self-supervised strategies in the left module, where additional training pairs are created from the partial inputs.

\vspace{-2mm}
\subsection{Completion Branch}
\label{Completion_Branch}
Our completion branch is a 
cascaded refinement architecture with three components: feature extraction, coarse reconstruction and dense reconstruction. 
Feature extraction is used to extract global features from the partial input. 
Coarse reconstruction generates sparse but complete point clouds from the global feature, and dense reconstruction synthesizes the dense and complete point clouds using both coarse outputs and partial inputs.

\subsubsection{Feature Extraction}
We use two stacked PointNet feature extraction architecture followed by max-pooling operations to extract global point features $f$ following PCN~\cite{yuan2018pcn}.
This simple design outperforms PointNet++~\cite{qi2017pointnet++} and DGCNN~\cite{dgcnn} since both PointNet++ and DGCNN consider the local feature aggregation, which is suboptimal due to the inaccurate selection of neighborhoods in the partial input. 

\begin{figure}
\centering
  \includegraphics[width=1\linewidth]{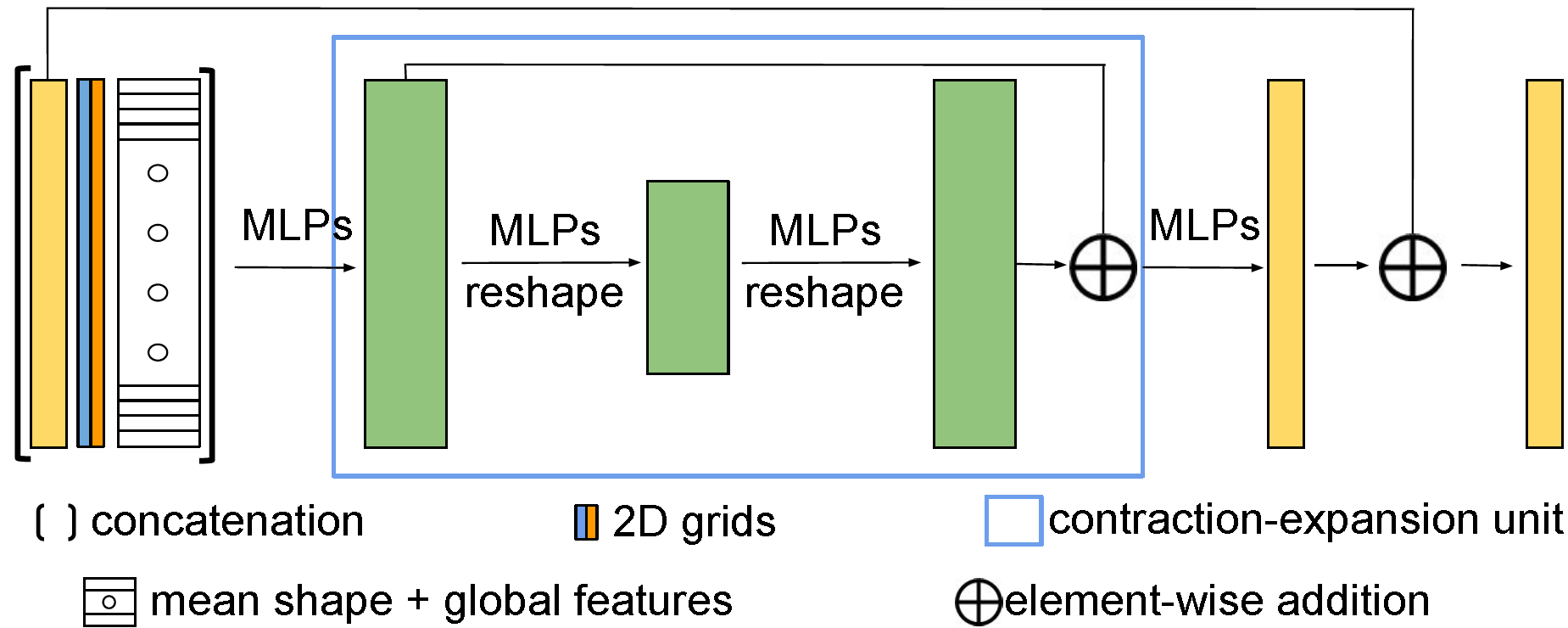}
  \vspace{-5mm}
  \caption{The architecture of the lifting module.}
  \label{lifting_unit}
  \vspace{-4mm}
\end{figure}

\begin{figure*}
\centering
  \includegraphics[width=1\linewidth]{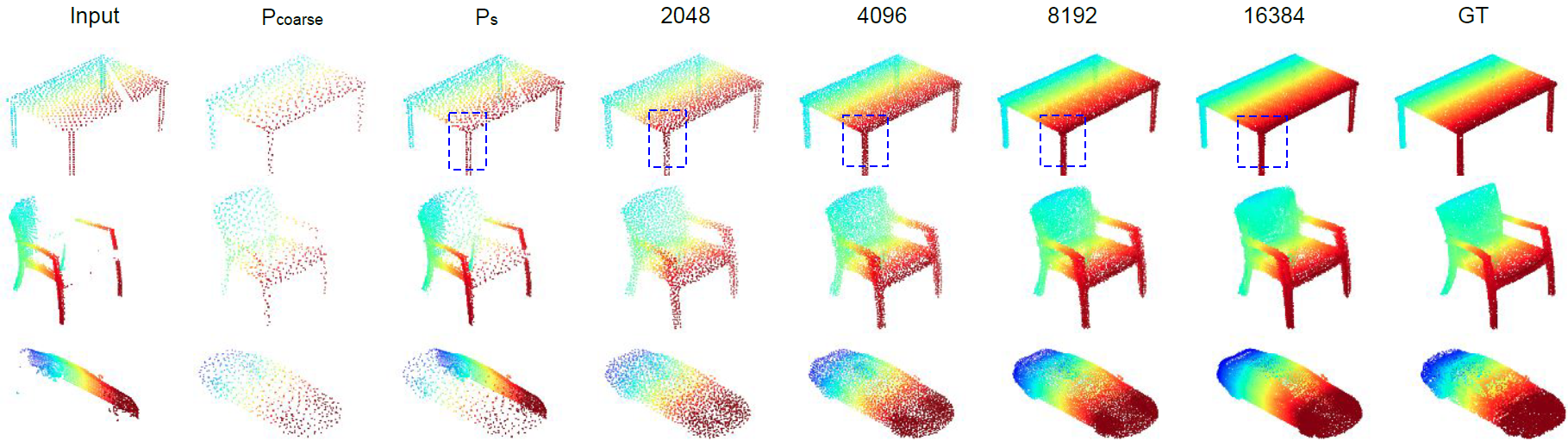}
  \vspace{-7mm}
  \caption{The results from the different stages of the completion process.}
  \label{iterative_results}
  \vspace{-6mm}
\end{figure*}

\subsubsection{Coarse Reconstruction}
Coarse reconstruction includes three fully connected layers (FC) mapping the global feature $f$ to the coarse point cloud $\text{P}_{\text{coarse}}$ with a size of $\text{N}_c\times 3$. 
Although multi-layer perceptions (MLPs) are able to generate the coarse outputs with fewer parameters, we argue that FC outperforms MLPs on synthesizing the global shape since FC considers long-range dependencies by connecting all the point features when generating new points.
The coarse outputs capture the complete object shapes, which are shown in the second column in Figure~\ref{iterative_results}.
Fine-grained object details are recovered in the dense reconstruction stage. 

\subsubsection{Dense Reconstruction}
Existing methods~\cite{yuan2018pcn,topnet2019,shu20193d} only consider the global features when generating dense and complete objects.
As can be seen in Figure~\ref{semi_pcn_qualitative}, both PCN~\cite{yuan2018pcn} and TopNet~\cite{topnet2019} fail to generate thin structures of 3D objects. 
The reason is that the output of the max-pooling layer $f$ mostly only contains the global shape information. It is challenging to recover local details from global information.
Considering this, we propose to use the partial input to preserve the object details together with the global embedding $f$ and the coarse output $\text{P}_{\text{coarse}}$ during the dense point reconstruction.
Specifically, 
we dynamically subsample $\text{N}_c\times 3$ points from the partial inputs $\text{P}$ and mirror the points regarding $xy$-plane~\cite{wang20193dn} since many man-made models show global reflection symmetry.
We then concatenate them with the coarse outputs $\text{P}_{\text{coarse}}$.
We denote the combined point sets as $\text{P}_S$ with a size of $2\text{N}_c\times 3$.
However, direct concatenation results in poor visual qualities because of the unevenly distributed points. 
We design a lifting module with a contraction-expansion unit to refine the point positions to alleviate this problem, as shown by the dense reconstruction after one iteration (the 4th column of Figure~\ref{iterative_results}).

\noindent\textbf{Remarks:} The mirror operation is used to provide the initial point coordinates, which are further refined by the lifting module. Nonetheless, this does not limit our method to only symmetric objects. We show that our network can handle non-symmetric objects in the supplementary materials.

The architecture of the lifting module is shown in Figure~\ref{lifting_unit}.
The input is the concatenation of the upsampled points, mean shape priors and the global feature $f$.
The point clouds are upsampled by the folding mechanism~\cite{yang2018foldingnet}, where every point is repeated two times and concatenated with a unique 2D grid vector sampled from a uniform distribution to increase the variation among the duplicated points.
We incorporate the mean shape prior to alleviate the domain gap of point features between the incomplete and complete point clouds. 
Subsequently, we feed the input to the contraction-expansion unit to consolidate the local and global features.
Point features are refined by first compressed into smaller sizes, and then restored to their original shapes in the contraction-expansion unit. 
Finally, our lifting module predicts point feature residuals 
instead of the final output since deep neural networks are better at predicting residuals~\cite{wang2018pixel2mesh}.

Given that directly synthesizing the dense and complete points by one single step 
gives rise to inferior outputs and also 
requires a large number of parameters~\cite{topnet2019}, we generate the dense and complete point clouds by iterating the lifting module several times.
Note that the parameters are shared among each iteration to reduce the model size. 
Specifically, the synthesis begins at generating lower resolution points ($\text{N}$=2,048), and points with higher resolutions $\text{N}=\{4,096, 8,192, 16,384\}$ are progressively obtained after $\{2, 3, 4\}$ iterations, respectively. 
Some results of different iterations are shown in Figure~\ref{iterative_results}.
Moreover, the object details are gradually polished and refined after more iterations, \emph{e.g.} the chair's legs in the first row (blue box).

\subsection{Partial Reconstruction Branch}
\label{Partial_Reconstruction}
As illustrated in the right bottom branch of Figure~\ref{whole_network}, we design the other partial reconstruction branch to train simultaneously with the completion branch. The partial reconstruction branch has an auto-encoder structure that aims to reconstruct the original partial input. The feature extractor of two branches shares the same weights during training. This ensures that it can learn features for both shape completion and partial reconstruction, which helps the network to keep the local details from the partial inputs in the feature space in the shape completion branch.

\subsection{Self-supervised Strategies}
\label{ss_strategies}
Given that large amounts of ground truth data are not always available 
for training, we propose two strategies to train the completion network with a self-supervised training scheme.
We first propose a resampling strategy to create training pairs from the partial inputs. 
The generated data that are more incomplete than the original training data induces our network to generate more complete data. 
We then make use of a point MixUp strategy~\cite{achituve2021self} to create additional training pairs for self-supervised learning.
Note that the created training pairs by the two strategies can be integrated with the original training data in the standard supervised learning setting. 
To this end, we experimentally show that the proposed strategies can also improve the performance of supervised settings. 

\subsubsection{Resampling}
\label{resampling}
We randomly remove a 
region from the original partial inputs $\text{P}$ to create another set of more incomplete point clouds $\text{P}'$. This set of point clouds and 
the original incomplete ones form our newly created training pairs, which we denote as $\text{S}' = \{\text{P}', \text{P}\}$. Our approach can be completely self-supervised when only trained on $\text{S}'$. Moreover, the combination of the original data $\text{S} = \{\text{P}, \text{Q}^{\prime}\}$ and $\text{S'}$ can enhance our approach by improving the generalization capacity, where $\text{Q}^{\prime}$ is the original ground truth. The intuition is that $\text{S}'$ consists of various of more incomplete point clouds with different missing regions. This diversity of $\text{S}'$ enforces our network to learn how to generate more complete shapes instead of memorizing the complete ground truth shapes in $\text{S}$.

\subsubsection{MixUp}
\label{MixUp}
\begin{figure}[!htb]
	\includegraphics[width=\linewidth]{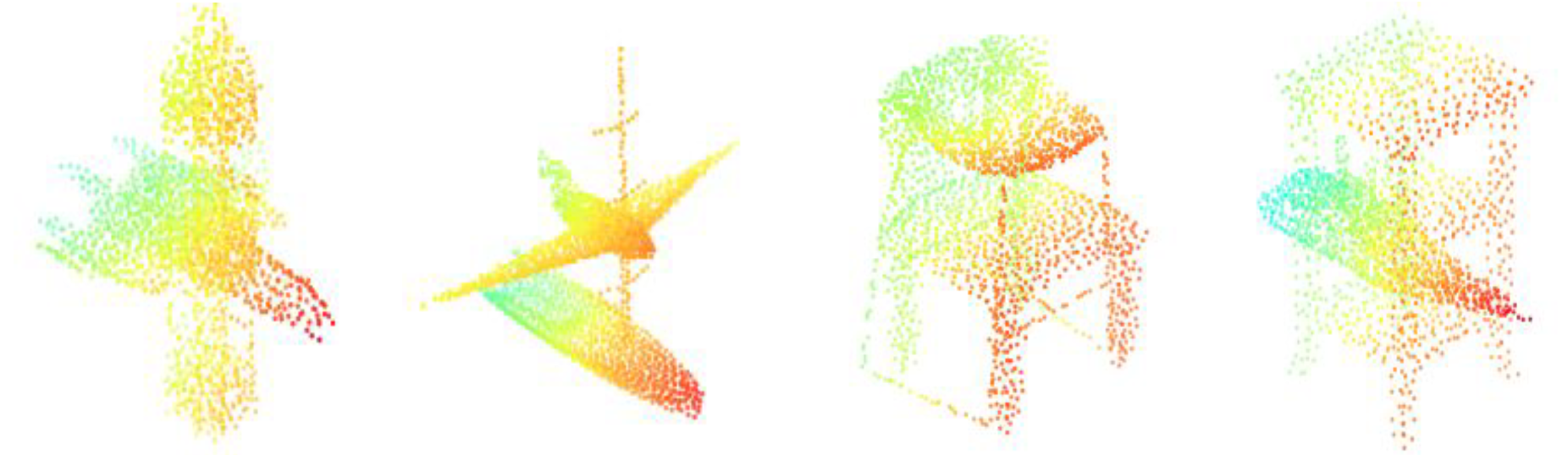}
	\vspace{-5mm}
	\caption{Some point cloud examples obtained by the MixUp operation.}
	\vspace{-5mm}
	\label{mixup_pic}
\end{figure}

The MixUp strategy~\cite{zhang2018mixup,achituve2021self} is initially proposed to improve the generalization of deep networks on classifications. We adapt it to create more training pairs in our 3D shape completion task. Specifically, given a pair of training data $(p_1^{\prime}, p_1)$ and $(p_2^{\prime}, p_2)$, we generate a new pair according to:
\begin{align}
    \begin{split}
    p^{\prime}&=\gamma p_1^{\prime} \oplus (1-\gamma)p_2^{\prime},\\
    p&=\gamma p_1 \oplus (1-\gamma)p_2, 
    \end{split}
\end{align}
where $\gamma \in [0, 1]$ is sampled from a $Beta$ distribution. The number of points sampled from each example are determined by $\gamma$, \emph{e.g.} $40 \%$ of the points are from $p_1^{\prime}$ and $60 \%$ are from $p_2^{\prime}$ when $\gamma=0.4$. 
$p_1^{\prime}$ and $p_2^{\prime}$ are the more incomplete point sets, and $p_1$ and $p_2$ are the original partial point sets. 
A set of points from $p_1^{\prime}$ and $p_2^{\prime}$ are sampled and concatenated to obtain the merged samples. Several merged examples from the complete point sets are shown in Figure~\ref{mixup_pic}. 
Although the created objects are not realistic, they are helpful to prevent the network from overfitting to the training data, especially when there is limited diversity in the training data.

\subsection{Optimization}
We make use of the Chamfer Distance (CD)~\cite{fan2017point} as our reconstruction loss.
CD computes the average closest point distance between two point clouds $\text{X}$ and $\text{Y}$, \emph{i.e.}:
\begin{align}
    \begin{split}
    \text{CD}(\text{X},\text{Y})&=\mathcal{L}_{\text{X},\text{Y}}+ \mathcal{L}_{\text{Y}, \text{X}},~ \text{where}\\
    \mathcal{L}_{\text{X},\text{Y}}&=\frac{1}{|\text{X}|}\sum_{x\in \text{X}}  \min\limits_{y\in \text{Y}}||x-y||_2,~\text{and}\\
    \mathcal{L}_{\text{Y},\text{X}}&=\frac{1}{|\text{Y}|}\sum_{y\in \text{Y}}  \min\limits_{x\in \text{X}}||x-y||_2.\\
    \end{split}
\end{align}
There are two variants for CD, which we denote as CD-1 and CD-2. 
CD-2 differs from CD-1 by taking the square root of the Euclidean distance. 
We adopt $\text{CD}\text{-}\text{2}$ in all our experiments during training, and evaluate with CD-1 and CD-2 on different datasets for fair comparisons.

Our training loss comprises three components: two reconstruction losses to encourage the completed point cloud to be the same as the ground truth, a reconstruction loss on the partial inputs and an adversarial loss to penalize unrealistic outputs.

\myparagraph{Reconstruction Losses.}
Following PCN~\cite{yuan2018pcn}, we compute the CD losses for the coarse and dense outputs.
The reconstruction losses can then be expressed as:
\begin{align}
\mathcal{L}_{\text{rec}}=\text{CD}(\text{P}_{\text{coarse}},\text{Q}^{\prime})+\lambda_{f} \text{CD}(\text{Q},\text{Q}^{\prime})+\lambda_{ae} \text{CD}(\text{\^{P}},\text{P}),
\end{align}
where $\text{P}_{\text{coarse}}$, $\text{Q}$, $\text{\^{P}}$ and $\text{Q}^{\prime}$ are the coarse output, dense output, the reconstructed partial point cloud and the ground truth, respectively. 
$\lambda_f$ is the weight for the CD of dense output, and $\lambda_{ae}$ is the weight for the self-supervised auto-encoder.

\myparagraph{Adversarial Loss.}
We adopt the stable and efficient objective function of LS-GAN~\cite{mao2017least} as our adversarial losses. Specifically, the adversarial losses for the generator and discriminator are:
\begin{align}
    \mathcal{L}_{\text{GAN}}(G) = \frac{1}{2}[D(\text{Q})-1]^2,
\end{align}
\begin{align}
    \mathcal{L}_{\text{GAN}}(D) = \frac{1}{2}[D(\text{Q})^2 +(D(\text{Q}^{\prime})-1)^2 ].
\end{align}
Our cascaded refinement sub-network is treated as the generator $G$. More architecture details of the discriminator $D$ are in the supplementary materials.

\myparagraph{Overall Loss.}
Our overall loss is the weighted sum of the reconstruction losses and the adversarial loss:
\begin{align}
    \mathcal{L}=\lambda \mathcal{L}_{\text{GAN}} + \beta \mathcal{L}_{\text{rec}},
\end{align}
where $\lambda$ and $\beta$ are the weights for the adversarial loss and reconstruction losses, respectively.
During training, $G$ and $D$ are optimized alternatively.

\begin{table*}[!htbp]
\centering
\resizebox{0.9\textwidth}{!}{
\small
\begin{tabular}{l|c| c |c  |c  | c|  c|  c |c|c }
\hline
\multirow{2}{*}{Methods} & \multicolumn{9}{c}{Mean Chamfer Distance per point ($10^{-3}$)} \\
\cline{2-10}
{}& Avg & Airplane & Cabinet & Car & Chair & Lamp & Sofa & Table & Vessel \\
\hline\hline
3D-EPN\cite{dai2017shape} &20.147 &13.161 &21.803 &20.306 &18.813 &25.746 &21.089 &21.716 &18.543  \\
FoldingNet~\cite{yang2018foldingnet} & 14.31 & 9.49 & 15.80 & 12.61 & 15.55 & 16.41 & 15.97 & 13.65 & 14.99 \\
AtlasNet~\cite{vakalopoulou2018atlasnet} & 10.85 & 6.37 & 11.94 & 10.10 & 12.06 & 12.37 & 12.99 & 10.33 & 10.61 \\
CRN (US) &11.980 &6.437 &15.093 &13.755 &12.366 &11.380 &14.703 &11.777 &10.331 \\
PCN-FC\cite{yuan2018pcn} &9.799 &5.698 &11.023 &8.775 &10.969 &11.131 &11.756 &9.320 &9.720 \\
PCN\cite{yuan2018pcn} &9.636 &5.502 &10.625 &8.696 &10.998 &11.339 &11.676 &8.590 &9.665 \\
TopNet~\cite{topnet2019} &9.890 &6.235 &11.628 &9.833 &11.498 &9.366 &12.347 &9.362 &8.851 \\
CRN (CD-1) &8.868 &5.065 & 10.383 & 8.650 & 9.830 & 9.830 & 11.002 & 8.253 & 8.335\\
GRNet~\cite{xie2020grnet} &8.828 &6.450 &10.373 &9.447 &9.408 &7.955 &10.512 &8.444 &8.039 \\
PMP-Net~\cite{wen2021pmp} &8.66 &5.50 &11.10 &9.62 &9.47 &\textbf{6.89} &10.74 &8.77 &\textbf{7.19} \\
CRN (CD-2) &8.505 &\textbf{4.794} &9.968 &\textbf{8.311} &9.492 &8.940 &10.685 &7.805 &8.045 \\
SCRN &\textbf{8.291} &4.795 &\textbf{9.935} &9.313 &\textbf{8.784} &8.657 &\textbf{9.737} &\textbf{7.201} &7.908 \\
\hline
\end{tabular}
}
\vspace{-2mm}
\caption{Quantitative comparison for point cloud completion on eight categories of objects from the PCN dataset. 
}
\vspace{-4mm}
\label{quantative_seen}
\end{table*}

\section{Experiments}
We compare our approach with several existing methods, which include 3D-EPN~\cite{dai2017shape}, FoldingNet~\cite{yang2018foldingnet}, AtlasNet~\cite{vakalopoulou2018atlasnet}, PCN~\cite{yuan2018pcn}, TopNet~\cite{topnet2019}, SoftPoolNet~\cite{wang2020softpoolnet}, SA-Net~\cite{wen2020point}, GRNet~\cite{xie2020grnet} and PMP-Net~\cite{wen2021pmp}.

\subsection{Evaluation Metrics}
We use the following evaluation metrics to quantitatively evaluate our results: a) CD, FPD, EMD, accuracy, completeness and F-score~\cite{Chen2020Unpaired,tatarchenko2019single} are used to evaluate the performances of synthetic datasets; b) Fidelity and plausibility are used to evaluate the performances of real-world datasets. 

\myparagraph{CD.}
We use CD-2 for the experiments in Section~\ref{exp_pcn}, and CD-1 in the remaining experiments for fair comparisons with PCN~\cite{yuan2018pcn} and TopNet~\cite{topnet2019}.

\myparagraph{FPD.}
Fr$\acute{\text{e}}$chet Point Cloud Distance (FPD)~\cite{shu20193d} is used to calculate the 2-Wasserstein distance between the real and fake Gaussian measurements in the feature spaces of the point sets, \emph{i.e.}:
\begin{equation}
    \begin{split}
    \text{FPD}(\text{X},\text{Y})=\|\text{m}_{\text{X}}-\text{m}_\text{Y}\|_2^2 +
    \text{Tr}(\Sigma_{\text{X}}+\Sigma_{\text{Y}} -2(\Sigma_{\text{X}}\Sigma_\text{Y})^{\frac{1}{2}}),
    \end{split}
\end{equation}
where m is the mean vector and $\Sigma$ is the covariance matrix of the points. Tr($A$) is the trace of matrix $A$. 

\myparagraph{EMD.}
Earth Mover Distance (EMD)~\cite{fan2017point} measures the average distance between corresponding points obtained from a bijection $\Phi:X\xrightarrow{}Y$. It is used to evaluate the uniformity of the generated point clouds.
\begin{align}
    \text{EMD}(X,Y)=\min\limits_{\Phi:X\xrightarrow{}Y}\frac{1}{|X|}\sum\limits_{x\in X}||x-\Phi(x)||_2.
\end{align}

\myparagraph{Accuracy.}
Accuracy measures the fraction of points in the output that are matched with the ground truth~\cite{Chen2020Unpaired}. Specifically, 
we compute $D(p,Q_{GT})=min(||p-q||,q\in Q_{GT})$ for each point $p$ in the output, and take the point as a correct match if $D(p,Q_{GT})<0.03$. 

\myparagraph{Completeness.}
Similar to accuracy, completeness reports the fraction of points in the ground truth that are within a distance threshold to any point in the output~\cite{Chen2020Unpaired}. 

\myparagraph{F-score.}
F-score is calculated as the harmonic average of the accuracy and completeness~\cite{Chen2020Unpaired,tatarchenko2019single}.

\myparagraph{Fidelity.}
Fidelity measures how well the inputs are preserved in the outputs~\cite{yuan2018pcn}. 

\myparagraph{Plausibility.}
Plausibility~\cite{Chen2020Unpaired} is evaluated as the classification accuracy in percentage by a pre-trained PointNet~\cite{qi2017pointnet} model. 

\begin{table}[]
\centering
\begin{tabular}{l c c c c c}
\hline
 & PCN~\cite{yuan2018pcn} &TopNet~\cite{topnet2019}  & GRNet~\cite{xie2020grnet} & CRN & SCRN \\
\hline
Acc. &0.946 &0.971 &\textbf{0.976} &0.959 &0.961 \\
Comp. &0.976 &0.967 &0.981 &0.980 &\textbf{0.984} \\
F-score &0.961 &0.969 &\textbf{0.979} &0.970 &0.973 \\
\hline
\end{tabular}
\vspace{-2mm}
\caption{Quantitative comparisons on the PCN dataset. Acc. and Comp. represent accuracy and completeness, respectively.}
\vspace{-4mm}
\label{quantitaitve_pcn_f_score}
\end{table}
\subsection{Datasets}
We test our method on both synthetic and real-world datasets. 
For the synthetic datasets, there are two types of methods to obtain the partial points: \textbf{1)} generate the incomplete points by back-projecting 2.5D depth images from a partial view into the 3D space~\cite{yuan2018pcn,topnet2019,wang2020cascaded}; \textbf{2)} randomly select a view point as the center and remove the points within a certain radius from the complete point sets~\cite{huang2020pf,sarmad2019rl,richard2020kaplan}.

Following method \textbf{1)}, we first evaluate completion performances on the PCN dataset~\cite{yuan2018pcn} and Completion3D benchmark~\footnote{https://completion3d.stanford.edu/results} for fair comparisons. 
The PCN dataset is generated from the ShapeNet dataset~\cite{wu20153d}. It has 30,974 objects, and eight categories: airplane, cabinet, car, chair, lamp, sofa, table and vessel.
Each incomplete point cloud consists of 2,048 points, and each complete point clouds consists of 16,384 points uniformly sampled from the mesh model. 
Completion3D~\cite{topnet2019} has the same categories with the PCN dataset. 
We follow the settings of training, validation and testing splits in PCN and Compeltion3D for fair comparisons. 

To test the generalization ability on more categories and the capability of reconstructing more complex structures, we also evaluate our method on the ModelNet dataset~\cite{wu20153d} which obtains partial points using method \textbf{2)}. 
This dataset has ten categories: bathtub, bed, bookshelf, cabinet, chair, lamp, monitor, plant, sofa and table.
The ground truth point clouds are uniformly sampled from the meshes.
Each complete point cloud consists of 2,048 points, and each partial input is obtained by randomly removing 512 points from the complete points.
The dataset has 4,183 training samples and 856 test samples. 

To verify the robustness of our method, we also test on the car category from the KITTI dataset~\cite{geiger2013vision} and the chair category from the MatterPort~\cite{Matterport3D} dataset.

\begin{table*}[!htbp]
\centering
\resizebox{0.85\textwidth}{!}{
\small
\begin{tabular}{l|c| c |c  |c  | c|  c|  c |c|c }
\hline
\multirow{2}{*}{Methods} & \multicolumn{9}{c}{Mean Chamfer Distance per point ($10^{-4}$)} \\
\cline{2-10}
{}& Avg & Airplane & Cabinet & Car & Chair & Lamp & Sofa & Table & Vessel \\
\hline\hline
FoldingNet\cite{yang2018foldingnet} &19.07 &12.83 &23.01 &14.88 &25.69 &21.79 &21.31 &20.71 &11.51 \\
PCN\cite{yuan2018pcn} &18.22 &9.79 &22.70 &12.43 &25.14 &22.72 &20.26 &20.27 &11.73 \\
PointSetVoting\cite{zhang2021point} &18.18 &6.88 &21.18 &15.78 &22.54 &18.78 &28.39 &19.96 &11.16 \\
AtlasNet\cite{groueix2018} &17.77 &10.36 &23.40 &13.40 &24.16 &20.24 &20.82 &17.52 &11.62 \\
TopNet~\cite{topnet2019} &14.25 &7.32 &18.77 &12.88 &19.82 &14.60 &16.29 &14.89 &8.82 \\
SoftPoolNet\cite{wang2020softpoolnet} &11.90 &4.89 &18.86 &10.17 &15.22 &12.34 &14.87 &11.84 &6.48 \\
SA-Net\cite{wen2020point} &11.22 &5.27 &14.45 &7.78 &13.67 &13.53 &14.22 &11.75 &8.84 \\
GRNet\cite{xie2020grnet} &10.64 &6.13 &16.90 &8.27 &12.23 &10.22 &14.93 &10.08 &5.86 \\
PMP-Net\cite{wen2021pmp} &9.23 &3.99 &14.70 &8.55 &10.21 &\textbf{9.27} &\textbf{12.43} &\textbf{8.51} &\textbf{5.77} \\
CRN &9.21 &3.38 &13.17 &8.31 &10.62 &10.00 &12.86 &9.16 &5.80 \\
SCRN &\textbf{9.13} &\textbf{3.35} &\textbf{12.81} &\textbf{7.78} &\textbf{9.88} &10.12 &12.95 &9.77 &6.10 \\
\hline
\end{tabular}
}
\vspace{-2mm}
\caption{Quantitative comparison for point cloud completion on eight categories objects of Completion3D benchmark.}
\vspace{-2mm}
\label{quantitative_topnet}
\end{table*}
\begin{figure*}
\centering
  \includegraphics[width=1\linewidth]{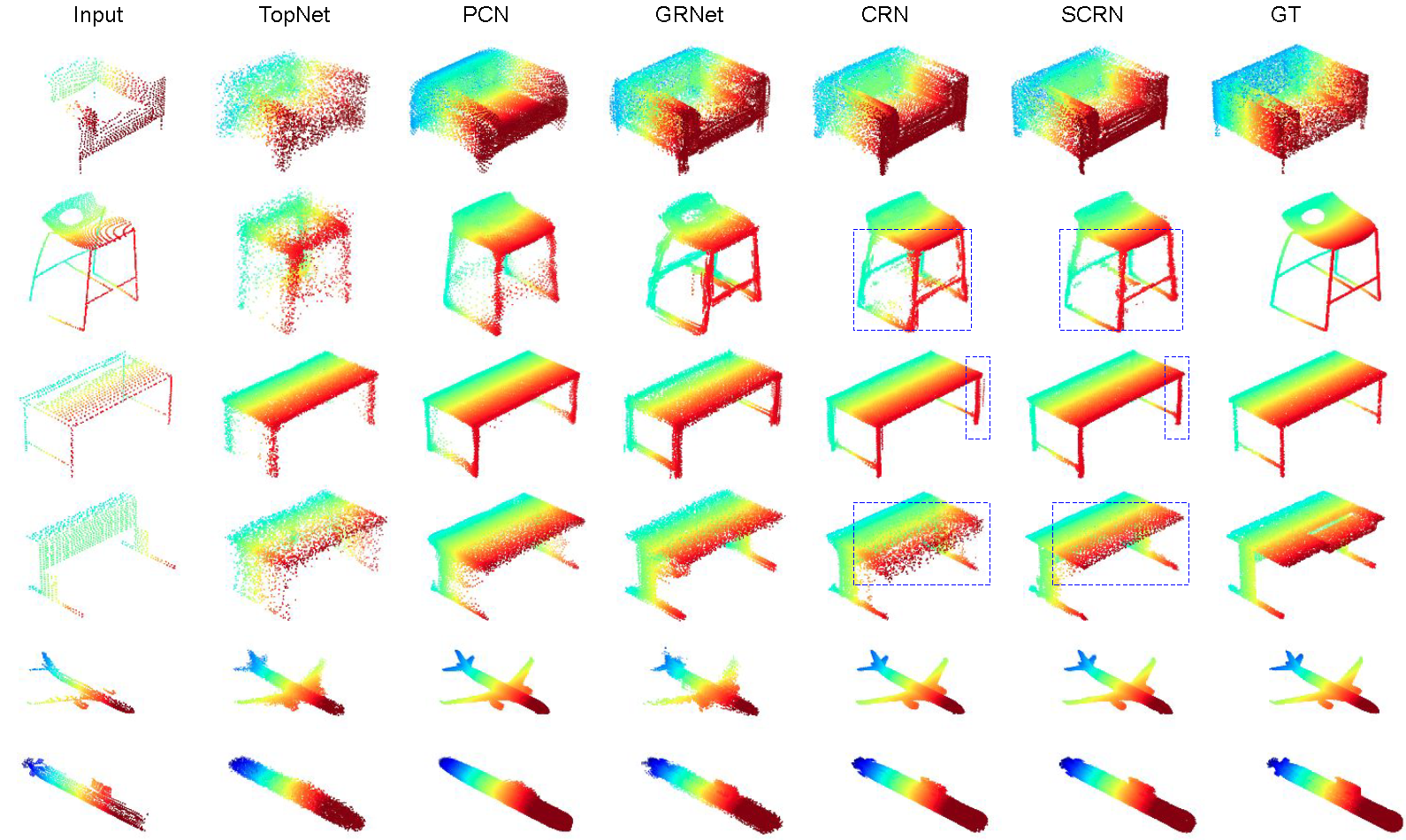}
  \vspace{-2mm}
  \caption{Qualitative comparison on the PCN dataset. Point resolutions for the output and ground truth are 16,384.}
  \vspace{-5mm}
  \label{semi_pcn_qualitative}
\end{figure*}

\subsection{Implementation Details}
We train our models using the Adam~\cite{kingma2014adam} optimizer.
We adopt the two time-scale update rule (TTUR)~\cite{heusel2017gans} and set the learning rates for the generator and discriminator as 0.0001 and 0.00005, respectively. 
The learning rates are decayed by 0.7 after 
every 40 epochs and clipped by $10^{-6}$ for the PCN dataset. 
$\lambda$, $\lambda_{ae}$ and $\beta$ are set to 1, 100 and 200, respectively.  
$\lambda_\text{f}$ increases from 0.01 to 1 within the first 50,000 iterations for the PCN dataset, and is set to 1 for the other datasets. We set $\text{N}_c=512$ for the coarse output.
We train one single model for all object categories.
The mean shape vectors are obtained by a pre-trained PointNet auto-encoder\footnote{https://github.com/charlesq34/pointnet-autoencoder} on all object categories following~\cite{kanazawa2018end}. We take mean values of latent embeddings from all instances within that category as our mean shape vectors for each object class. 

We use CRN to denote the standard supervised learning only with our completion branch. 
CRN+US denotes self-supervised learning, and SCRN denotes hybrid training of the supervised and self-supervised strategies.
Both CRN+US and SCRN are trained with our two-branch network.
Refer to the source code for CRN at \url{https://github.com/xiaogangw/cascaded-point-completion} for more implementation details.
More experimental results are in the supplementary materials.

\begin{figure*}
	\centering
	\includegraphics[width=0.95\linewidth]{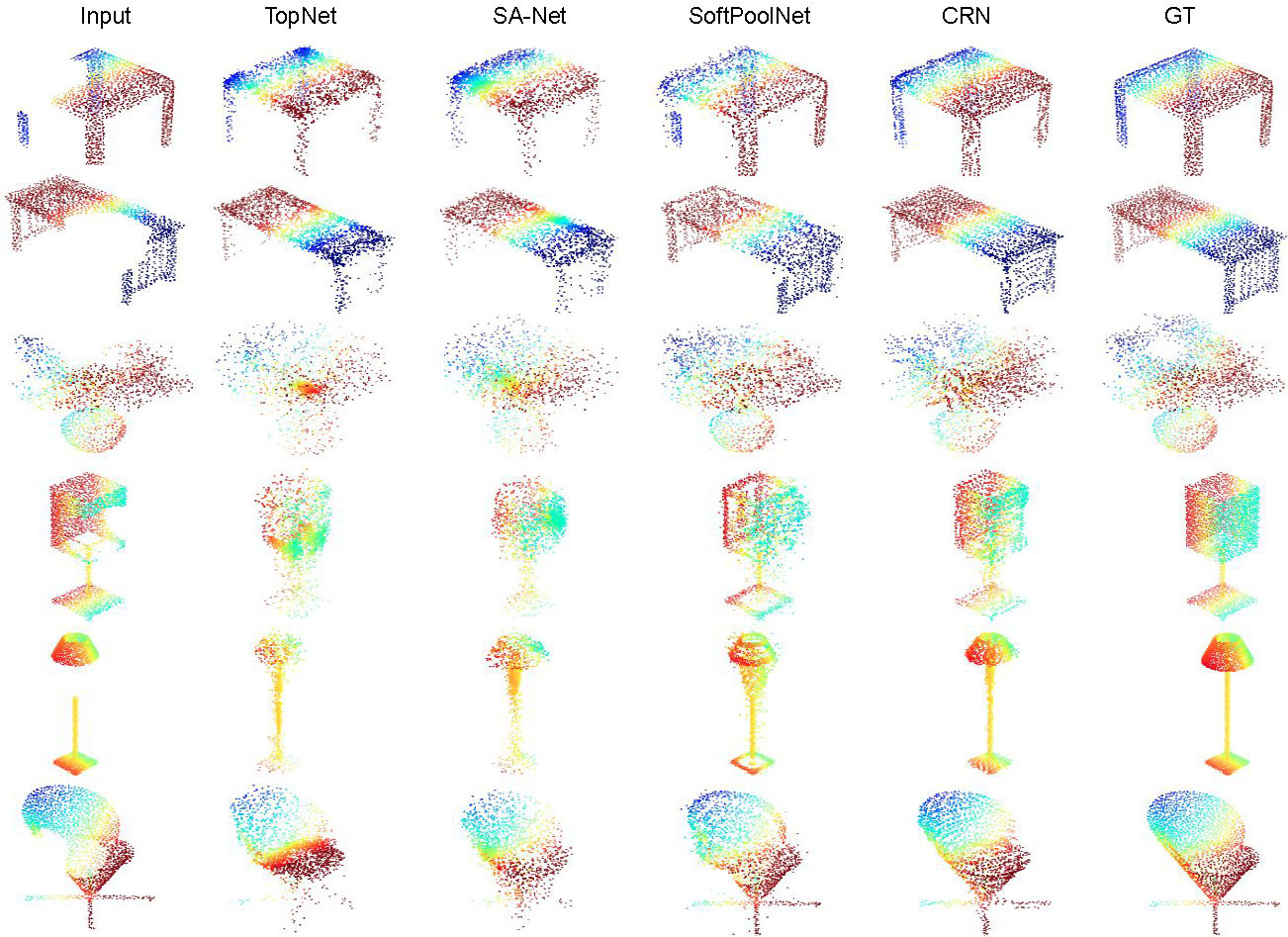}
	\caption{Qualitative comparison on the ModelNet dataset. The resolution for partial points and ground truth are 2,048.}
	\vspace{-2mm}
	\label{topnet_pcn_condition10_comparison_modelnet}
\end{figure*}

\subsection{Point Completion on the PCN dataset}\label{exp_pcn}
Quantitative and qualitative results are shown in Tables~\ref{quantative_seen},~\ref{quantitaitve_pcn_f_score} and Figure~\ref{semi_pcn_qualitative}, respectively. 
The resolutions of the output and the ground truth are set to 16,384. 
Some results are directly cited from PMP-Net~\cite{wen2021pmp}, and the qualitative results of GRNet are generated by their released model. 
The quantitative results in Table~\ref{quantative_seen} show that our cascaded refinement network (CRN) obtains the best performance on the average CD error compared to existing state-of-the-art methods.
Our proposed approach SCRN further improves the performance. 
Although the PCN dataset has a large amount of training data (231,792 training samples compared to 4,183 training data for ModelNet), SCRN is able to obtain 3$\%$ relative improvements compared to CRN with our proposed techniques. 
This verifies that the auto-encoder together with our self-supervised strategies are able to obtain more accurate global shapes and finer local structures.
Furthermore, 
we use data created from the self-supervised strategies denoted as CRN (US) to evaluate our method on self-supervision.  
Although the performance is worse than our supervised approach (11.980 vs 8.505), our self-supervised results are better than the supervised methods 3D-EPN~\cite{dai2017shape} and FoldingNet~\cite{yang2018foldingnet}. This verifies that our proposed self-supervised strategies can generate reasonable objects without using fully complete ground truth. 
We adopt EMD, accuracy, completeness and F-score~\cite{tatarchenko2019single} as additional evaluation metrics to evaluate the uniformity 
of the point distribution and the reconstruction quality shown in Table~\ref{quantitaitve_pcn_f_score}. The results show that we achieve the best performance on completeness and are comparable with GRNet on accuracy and F-score. This verifies that our method is able to reconstruct high-quality objects. 

We also compare the qualitative results with existing works in Figure~\ref{semi_pcn_qualitative}. 
Our methods CRN and SCRN generate more accurate object shapes than the other approaches, especially for the thinner parts, \emph{e.g.} the chair and table legs.
Moreover, our improved method SCRN generates better object details than our method CRN. This is verified by the reconstructed table planes in the 4th row (blue box). 
The 
noisy points are also successfully removed in the results of SCRN as shown in the chair results of the 2nd and 3rd rows (blue box). This further verifies the effectiveness of our proposed self-supervised strategies.

\subsection{Point Completion on the Completion3D Benchmark}\label{exp_topnet}
We compare the quantitative results on the Completion3D benchmark in Table~\ref{quantitative_topnet}, where our method CRN achieves better performances in terms of the average CD error across all categories compared to the previous state-of-the-art algorithms. 
We further boost the performance with our improved architecture SCRN. 
The superior performances of SCRN over CRN show that our self-supervised strategies are complementary to the supervised architectures, and the partial reconstruction can be integrated with the completion pipeline seamlessly to preserve local details from the partial inputs. This also verifies that the created training pairs from the resampling and MixUp strategies prevent our method from overfitting to the training data.

\begin{table*}[!htbp]
\centering
\resizebox{0.95\textwidth}{!}{
\small
\begin{tabular}{l|c| c |c  |c  | c|  c|  c |c|c |c|c }
\hline
\multirow{2}{*}{Methods} & \multicolumn{9}{c}{Mean Chamfer Distance per point ($10^{-3}$)} \\
\cline{2-12}
{}& Avg & Bathtub & Bad & Bookshelf & Cabinet & Chair & Lamp & Monitor & Plant & Sofa & Table \\
\hline\hline
TopNet~\cite{topnet2019} &4.172 &3.767 &2.829 &3.745 &4.208 &3.247 &17.862 &3.295 &7.963 &3.208 &2.345 \\
SA-Net~\cite{wen2020point} &4.541 &3.881 &2.824 &4.654 &4.643 &3.521 &18.159 &3.371 &8.838 &3.321 & 2.770\\
SoftPoolNet~\cite{wang2020softpoolnet} &3.898 &3.944 &2.892 &3.754 &4.839 &2.889 &16.033 &3.205 &5.427 &3.241 &2.623 \\
CRN (US) &3.491 &3.997 &3.120 &3.056 &4.194 &2.356 &4.796 &3.433 &5.352 &3.645 &2.358 \\
CRN &2.606 &\textbf{2.329} &1.747 &2.311 &2.828 &1.618 &8.258 &\textbf{2.120} &5.394  &\textbf{2.104} &1.766 \\
SCRN &\textbf{2.472} &2.409 &\textbf{1.717} &\textbf{2.258} &\textbf{2.804} &\textbf{1.607} &\textbf{6.133} &2.121 &\textbf{4.862} &2.112 &\textbf{1.645} \\
\hline
\end{tabular}
}
\vspace{-1mm}
\caption{Quantitative comparison for point cloud completion on ten categories objects from the ModelNet dataset. 
}
\vspace{-1mm}
\label{quantitative_model}
\end{table*}
\begin{table}[]
\centering
\begin{tabular}{l c c c c}
\hline
 & EMD & Accuracy &Completeness  & F-score \\
\hline
TopNet~\cite{topnet2019}  &0.160 & 0.561 &0.431 &0.487 \\
SA-Net~\cite{wen2020point}  &0.233 &0.537 &0.397 &0.456 \\
SoftPoolNet~\cite{wang2020softpoolnet}  &0.129 &0.705 &0.566 &0.628 \\
CRN &\textbf{0.123} &0.707 &\textbf{0.626} &\textbf{0.664} \\
SCRN &0.139 &\textbf{0.720} &0.613 &0.663 \\
\hline
\end{tabular}
\vspace{-1mm}
\caption{Quantitative comparisons on the ModelNet dataset.}
\label{quantitaitve_modelnet_f_score}
\end{table}

\begin{figure*}
	\includegraphics[width=\linewidth]{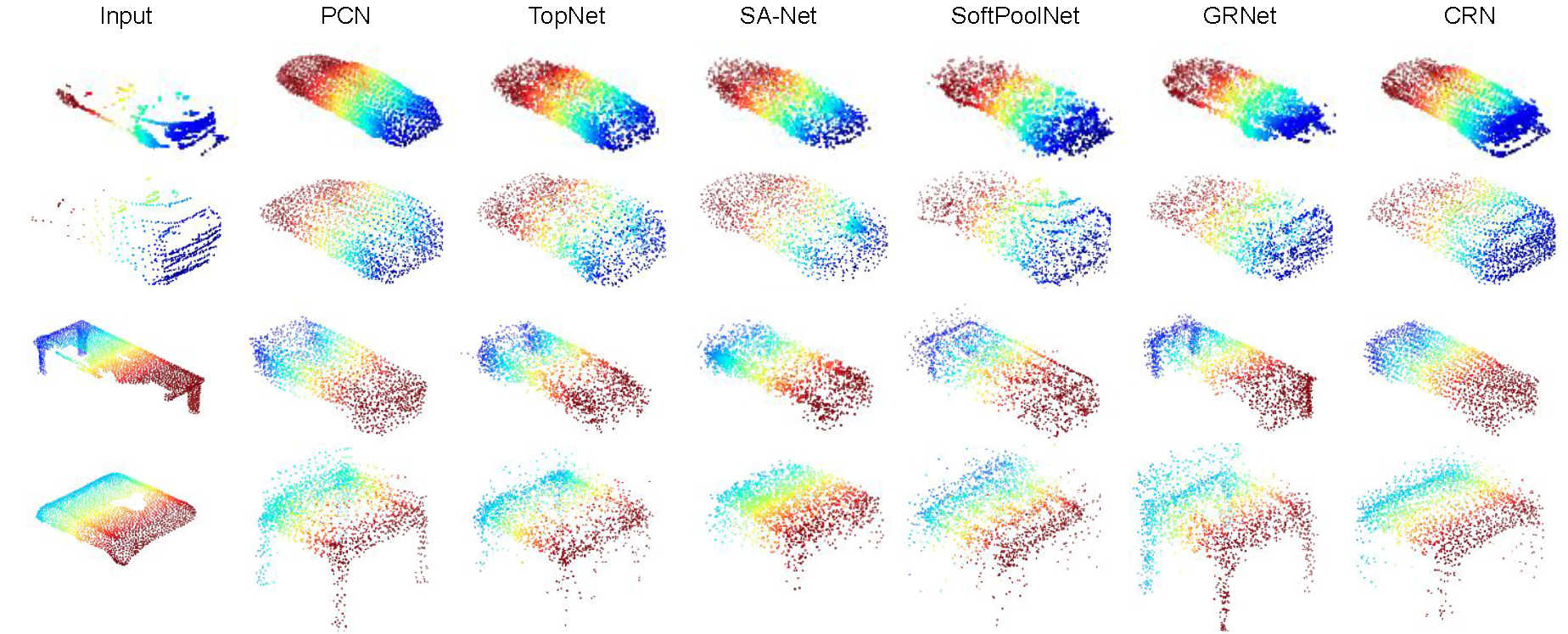}
	\vspace{-5mm}
	\caption{Qualitative results on the real-world dataset. The first two rows are the cars from the KITTI dataset, and the last two rows are the tables from the MatterPort dataset.}
	\label{qualitative_realdata}
	\vspace{-4mm}
\end{figure*}

\subsection{Point Completion on the ModelNet Dataset}
Quantitative and qualitative results on the ModelNet dataset~\cite{qin2019pointdan,achituve2021self} are shown in Tables~\ref{quantitative_model},~\ref{quantitaitve_modelnet_f_score} and Figure~\ref{topnet_pcn_condition10_comparison_modelnet}, respectively. 

The quantitative comparisons in Tables~\ref{quantitative_model} and~\ref{quantitaitve_modelnet_f_score} indicate that our method CRN achieves the best performance compared to existing approaches in different evaluation metrics, which verify the robustness and generalization ability of CRN. 
The qualitative results in Figure~\ref{topnet_pcn_condition10_comparison_modelnet} show that our method is able to generate the details not only included in the partial scan, but also for the missing parts, \emph{e.g.} the legs of tables and chairs. In addition, our generated shapes are more realistic compared to other methods, such as the results of lamp and plant.

\begin{table}[!htbp]
\centering
\small
\begin{tabular}{l|c| c |c  |c  | c}
\hline
\multirow{2}{*}{Methods} &  \multicolumn{5}{c}{Chamfer Distance ($10^{-3}$)}\\
\cline{2-6}
{}& 10\% & 30\% & 50\% & 70\% & 90\%\\
\hline\hline
PCN &6.520 &4.923 &4.280 &3.942 &3.661 \\
PCN (US)$^{\dagger}$ &\textbf{4.496} &\textbf{3.375} &\textbf{3.028} &\textbf{2.837} &\textbf{2.829} \\
Improv.($\%$) &31.0 &31.4 &29.3 &28.0 &22.7 \\
\hline\hline
CRN &4.291 &3.413 &2.981 &2.803 &2.669 \\
CRN (US)$^{\dagger}$ &\textbf{3.331} &\textbf{2.816} &\textbf{2.600} &\textbf{2.494} &\textbf{2.433}\\
Improv.($\%$) &22.4 &17.5 &12.8 &11.0 &8.8 \\
\hline
\end{tabular}
\vspace{-1.5mm}
\caption{Quantitative comparisons for semi-supervised learning on ten categories of objects on the ModelNet dataset. Results are represented by mean Chamfer Distance per point ($10^{-3}$). $^{\dagger}$ represents the training with both labeled and unlabeled data.}

\label{modelnet_semisupervised}
\vspace{-5mm}
\end{table}
To further verify the effectiveness of our self-supervised learning, we evaluate our approach on the self-supervised setting (CRN (US)) and the hybrid training.
We treat all the training examples as unlabeled data for the self-supervised experiments. 
We can obtain the conclusion that our method under self-supervised setting is capable of completing the objects without accessing any fully complete point sets. As the results shown in Table~\ref{quantitative_model}, our self-supervised result CRN (US) already outperforms other methods, and SCRN further improves the performances. This also verifies the superiority of our cascaded refinement method.
In the hybrid training shown in Table~\ref{modelnet_semisupervised}, we draw varying ratios of labeled data out of the original training set in different experiments, \emph{e.g.} 30$\%$ represents 30$\%$ of training data are original training pairs and 70$\%$ of training data are our created pairs. We denote them as semi-supervised settings and show the results in Table~\ref{modelnet_semisupervised}.
With the ratio of labeled data decreases, the margin between our method and the fully supervised methods increases. Given 10$\%$ labeled data, our self-supervised strategies improve around 31.0$\%$ and 22.4$\%$ over PCN and CRN on the ModelNet dataset, respectively. 
This indicates that our self-supervised strategies together with the auto-encoder sub-network improve the semi-supervised performances consistently. Moreover, our proposed strategies are compatible with different backbones and can improve the performances persistently.

\begin{figure}
\centering
    \vspace{-4mm}
  \includegraphics[width=1\linewidth]{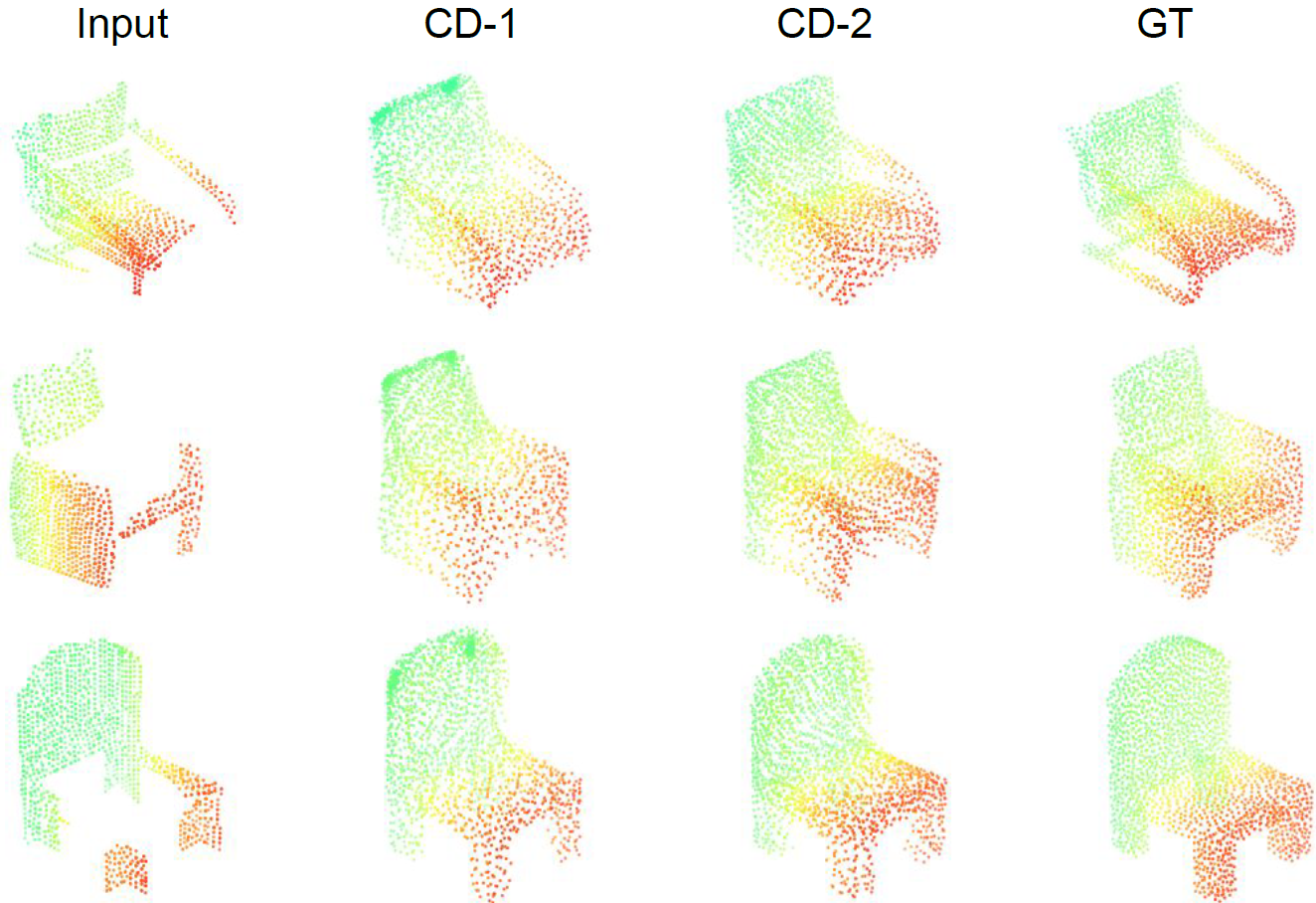}
  \vspace{-3mm}
  \caption{Qualitative comparison between two variants of Chamfer Distance.}
  \label{ablation_cds}
  \vspace{-3mm}
\end{figure}

\begin{table}[]
\centering
\begin{tabular}{l| c| c|| c| c}
\hline
\multirow{2}{*}{Methods} & \multicolumn{2}{c||}{Car} & \multicolumn{2}{c}{Table} \\
\cline{2-5}
& Fidelity & Plaus.($\%$) & Fidelity & Plaus.($\%$) \\
\hline
PCN~\cite{yuan2018pcn}  &1.221 &\textbf{59} &0.443 &45 \\
TopNet~\cite{topnet2019}  &1.530 &50  &0.393 &60 \\
SA-Net~\cite{wen2020point}  &0.681 &20 &0.286 &65 \\
SoftPoolNet~\cite{wang2020softpoolnet}  &0.313 &19  &0.349 &65\\
GRNet~\cite{xie2020grnet}  &0.132 &39  &0.191 &60\\
CRN  &\textbf{0.097} &57  &0.163 &65 \\
SCRN  &0.099 &57  &\textbf{0.144} &\textbf{65}\\
\hline
\end{tabular}
\caption{Quantitative comparisons on the real-world datasets. Plaus. represents for the classification accuracy produced by PointNet~\cite{qi2017pointnet}. Fidelity: the lower the better. Plaus.: the higher the better.}
\vspace{-3mm}
\label{quantitative_realworld}
\end{table}

\begin{table*}[!ht]
\centering
\resizebox{0.7\textwidth}{!}{
\begin{tabular}{c c c c   c  c  c  c  c  c}
\hline
 & PCN~\cite{yuan2018pcn} &TopNet~\cite{topnet2019}  & GRNet~\cite{xie2020grnet} & SoftPoolNet~\cite{wang2020softpoolnet}  & SA-Net~\cite{wen2020point} & CRN \\
\hline
Para. (M) &6.85 &9.96 &76.71 &160.43 &\textbf{1.67} &5.14\\
Time$^1$ (s) &0.22 &\textbf{0.18} &0.41 &0.42 &- &0.37\\
Time$^2$ (s) &0.050 &0.085 &0.20 &0.26 &0.13 &\textbf{0.050}\\
\hline
\end{tabular}
}
\vspace{-2mm}
\caption{Space and time comparisons of different methods. Time$^1$ and Time$^2$ represent the times of 16,384 and 2,048 resolutions, respectively.}
\vspace{-2mm}
\label{model_size}
\end{table*}
\begin{table}[!htbp]
\small
\centering
\begin{tabular}{l|c| c|c|c}
\hline
Method & BS & w/ Mir & w/ CE & w/ Dis\\
\hline
CD ($10^{-3}$) &6.523 &6.331 &6.243 &\textbf{6.181} \\
\hline
\end{tabular}
\vspace{-2mm}
\caption{Ablation studies on the bed category of the ModelNet dataset.}
\vspace{-2mm}
\label{ablation_modelnet_shapenet}
\end{table}

\begin{table}[!htbp]
\centering
\resizebox{0.4\textwidth}{!}{
\small
\begin{tabular}{c| c |c ||c }
\hline
Resampling & Mixup & Partial AE & ModelNet \\
\hline
\xmark &\xmark &\xmark  &6.887 \\
\xmark &\xmark &\cmark  &6.577 \\
\xmark &\cmark &\xmark  &6.050 \\
\cmark &\xmark &\xmark  &4.838 \\
\cmark &\cmark &\cmark  &\textbf{4.496} \\
\hline
\end{tabular}
}
\vspace{-2mm}
\caption{Ablation studies on the ModelNet dataset. Results are represented by mean Chamfer Distance per point (10$^{-3}$).}
\vspace{-2mm}
\label{ablation_results_self}
\end{table}

\subsection{Point Completion on the Real-world Datasets}
In addition to the synthetic dataset, we test on cars of the KITTI~\cite{geiger2013vision} dataset and tables of the MatterPort~\cite{Matterport3D} dataset as well to show the robustness of our methods. 
We take a sequence of Velodyne scans from the KITTI dataset following PCN to obtain 2,401 partial point clouds. 
20 examples for table category are extracted from the MatterPort~\cite{Matterport3D} dataset following Pcl2pcl~\cite{Chen2020Unpaired}.
Since existing methods cannot be trained on the real-world datasets due to the lack of ground truth, we directly take use of the pre-trained models on the Completion3D dataset for testing and show the qualitative results in Figure~\ref{qualitative_realdata}.
The average number of partial inputs in the KITTI dataset is 440, or even fewer than 10 points in some cases, which is much smaller than the average number of the synthetic ShapeNet dataset (larger than 1,000). 
Despite that, we obtain good results with high-fidelity outputs and hallucinate finer details at the same time compared to other approaches, \emph{e.g.} the rear parts of the cars in the first two rows of Figure~\ref{qualitative_realdata}, while other methods fail to generate the details and obtain low-quality results.
Our method also generates more precise table planes compared to other approaches, as can be seen from the last two rows of Figure~\ref{qualitative_realdata}. 

The quantitative results are shown in Table~\ref{quantitative_realworld}. Following PCN~\cite{yuan2018pcn} and Pcl2pcl~\cite{Chen2020Unpaired}, we adopt fidelity and plausibility as the evaluation metrics. 
Our methods CRN and SCRN achieve the best and second-best performance on fidelity for the car generation, respectively. Both CRN and SCRN obtain comparable classification accuracy to PCN. 
For the table generation, SCRN achieves the best performances on both fidelity and classification accuracy followed by CRN.
These quantitative results verify the robustness of our approach on the real-world datasets although no ground truth are available. 

\subsection{Time and Space Complexity Analysis}
We show the number of parameters and inference time of different methods in Table ~\ref{model_size}. 
The average inference time is computed by 5,000 forward steps on a Titan X GPU with two resolutional points: 2,048 and 16,384.
As shown in Table ~\ref{model_size}, our model consumes fewer parameters compared to other methods, especially the voxel-based method GRNet~\cite{xie2020grnet}.
The last row (Time$^2$) shows that PCN and our method are the most efficient algorithms in terms of inference time when the output resolution is 2,048. The 3rd row (Time$^1$) shows that we achieve comparable time efficiency with PCN and TopNet, and is better than GRNet and SoftPoolNet when output resolution is 16,384. This comparison verifies that our progressive refinement process does not consume extra time. Since SA-Net only generates the results with the resolution of 2,048, we only show their inference time for the resolution of 2,048.

\subsection{Ablation Studies}
We first evaluate the performance of our network trained with different variants of CDs on the PCN dataset.
The quantitative results in Table~\ref{quantative_seen} indicate that our network trained with CD-2 achieves better results.
Qualitative comparisons between CD-1 and CD-2 are shown in Figure~\ref{ablation_cds}.
Although both losses capture the general shapes with similar details of objects, the results of CD-2 are more evenly distributed compared to the results of CD-1. For example, more points are clustered on the top edges of chairs in the results of CD-1.

We also evaluate the effectiveness of 
the discriminator (Dis), the mirroring operation (Mir) and the contraction-expansion unit (CE) of our point completion network. We denote our method without discriminator, the mirroring operation or the contraction-expansion unit as the baseline (BS).
The quantitative comparison is shown in Table~\ref{ablation_modelnet_shapenet}.
All experiments are done on the point resolution of 2,048. We choose the bed category of the ModelNet dataset as an example.
The improved performances achieved by adding each component to the baseline model verify the contribution of each component.

We further analyze the effects of the partial reconstruction, resampling strategy and MixUp strategy on the ModelNet dataset with 10$\%$ labeled data.
The results are shown in Table~\ref{ablation_results_self}.
It is obvious that every component contributes to the whole pipeline since removing any part is detrimental to the performance.
Moreover, the superior results of SCRN against to CRN on the datasets of PCN (Table~\ref{quantative_seen}), Completion3D (Table~\ref{quantitative_topnet}) and ModelNet (Table~\ref{quantitative_model}) also verify the effectiveness of our proposed strategies.

\section{Conclusion}
In this work, we propose a two-branch shape completion network with self-supervision to generate complete points given the partial inputs. 
Our cascaded refinement generator exploits existing details from the partial inputs and synthesizes the missing parts with high quality. 
Furthermore, we propose two strategies to generate training pairs without the fully complete ground truth to enable our network to train with a self-supervised training scheme. 
Our results are evaluated on different completion datasets and achieved state-of-the-art performances.


%



\ifCLASSOPTIONcompsoc
  \section*{Acknowledgments}
\else
  \section*{Acknowledgment}
\fi
This research was supported in part by the Singapore Ministry of Education (MOE) Tier 1 grant R-252-000-A65-114, National University of Singapore Scholarship Funds, the National Research Foundation, Prime Ministers Office, Singapore, under its CREATE programme, Singapore-MIT Alliance for Research and Technology (SMART) Future Urban Mobility (FM) IRG, and the Singapore Agency for Science, Technology and Research (A*STAR) under its AME Programmatic Funding Scheme (Project $\#$A18A2b0046).





%


\bibliographystyle{IEEEtran}
\bibliography{IEEEabrv,IEEEexample}

\begin{thebibliography}{10}
\providecommand{\url}[1]{#1}
\csname url@samestyle\endcsname
\providecommand{\newblock}{\relax}
\providecommand{\bibinfo}[2]{#2}
\providecommand{\BIBentrySTDinterwordspacing}{\spaceskip=0pt\relax}
\providecommand{\BIBentryALTinterwordstretchfactor}{4}
\providecommand{\BIBentryALTinterwordspacing}{\spaceskip=\fontdimen2\font plus
\BIBentryALTinterwordstretchfactor\fontdimen3\font minus
  \fontdimen4\font\relax}
\providecommand{\BIBforeignlanguage}[2]{{%
\expandafter\ifx\csname l@#1\endcsname\relax
\typeout{** WARNING: IEEEtran.bst: No hyphenation pattern has been}%
\typeout{** loaded for the language `#1'. Using the pattern for}%
\typeout{** the default language instead.}%
\else
\language=\csname l@#1\endcsname
\fi
#2}}
\providecommand{\BIBdecl}{\relax}
\BIBdecl

\bibitem{qi2017pointnet}
C.~R. Qi, H.~Su, K.~Mo, and L.~J. Guibas, ``Pointnet: Deep learning on point
  sets for 3d classification and segmentation,'' in \emph{Proceedings of the
  IEEE Conference on Computer Vision and Pattern Recognition}, 2017, pp.
  652--660.

\bibitem{qi2017pointnet++}
C.~R. Qi, L.~Yi, H.~Su, and L.~J. Guibas, ``Pointnet++: Deep hierarchical
  feature learning on point sets in a metric space,'' in \emph{Advances in
  Neural Information Processing Systems}, 2017, pp. 5099--5108.

\bibitem{li2018so}
J.~Li, B.~M. Chen, and G.~Hee~Lee, ``So-net: Self-organizing network for point
  cloud analysis,'' in \emph{Proceedings of the IEEE conference on computer
  vision and pattern recognition}, 2018, pp. 9397--9406.

\bibitem{dgcnn}
Y.~Wang, Y.~Sun, Z.~Liu, S.~E. Sarma, M.~M. Bronstein, and J.~M. Solomon,
  ``Dynamic graph cnn for learning on point clouds,'' \emph{ACM Transactions on
  Graphics (TOG)}, 2019.

\bibitem{achituve2021self}
I.~Achituve, H.~Maron, and G.~Chechik, ``Self-supervised learning for domain
  adaptation on point clouds,'' in \emph{Proceedings of the IEEE/CVF Winter
  Conference on Applications of Computer Vision}, 2021, pp. 123--133.

\bibitem{lang2019pointpillars}
A.~H. Lang, S.~Vora, H.~Caesar, L.~Zhou, J.~Yang, and O.~Beijbom,
  ``Pointpillars: Fast encoders for object detection from point clouds,'' in
  \emph{Proceedings of the IEEE Conference on Computer Vision and Pattern
  Recognition}, 2019, pp. 12\,697--12\,705.

\bibitem{liang2018deep}
M.~Liang, B.~Yang, S.~Wang, and R.~Urtasun, ``Deep continuous fusion for
  multi-sensor 3d object detection,'' in \emph{Proceedings of the European
  Conference on Computer Vision (ECCV)}, 2018, pp. 641--656.

\bibitem{giancola2019leveraging}
S.~Giancola, J.~Zarzar, and B.~Ghanem, ``Leveraging shape completion for 3d
  siamese tracking,'' in \emph{Proceedings of the IEEE Conference on Computer
  Vision and Pattern Recognition}, 2019, pp. 1359--1368.

\bibitem{hou2019sis}
H.~Ji, A.~Dai, and M.~Nie{\ss}ner, ``3d-sis: 3d semantic instance segmentation
  of rgb-d scans,'' in \emph{Proceedings Computer Vision and Pattern
  Recognition (CVPR), IEEE}, 2019.

\bibitem{dai2018scancomplete}
A.~Dai, D.~Ritchie, M.~Bokeloh, S.~Reed, J.~Sturm, and M.~Nie{\ss}ner,
  ``Scancomplete: Large-scale scene completion and semantic segmentation for 3d
  scans,'' in \emph{Proceedings Computer Vision and Pattern Recognition (CVPR),
  IEEE}, 2018.

\bibitem{boud1999virtual}
A.~C. Boud, D.~J. Haniff, C.~Baber, and S.~Steiner, ``Virtual reality and
  augmented reality as a training tool for assembly tasks,'' in \emph{1999 IEEE
  International Conference on Information Visualization (Cat. No.
  PR00210)}.\hskip 1em plus 0.5em minus 0.4em\relax IEEE, 1999, pp. 32--36.

\bibitem{webster1996augmented}
A.~Webster, S.~Feiner, B.~MacIntyre, W.~Massie, and T.~Krueger, ``Augmented
  reality in architectural construction, inspection and renovation,'' in
  \emph{Proceedings ASCE Third Congress on Computing in Civil Engineering},
  vol.~1, 1996, p. 996.

\bibitem{yuan2018pcn}
W.~Yuan, T.~Khot, D.~Held, C.~Mertz, and M.~Hebert, ``Pcn: Point completion
  network,'' in \emph{2018 International Conference on 3D Vision}.\hskip 1em
  plus 0.5em minus 0.4em\relax IEEE, 2018, pp. 728--737.

\bibitem{topnet2019}
L.~P. Tchapmi, V.~Kosaraju, S.~H. Rezatofighi, I.~Reid, and S.~Savarese,
  ``Topnet: Structural point cloud decoder,'' in \emph{Proceedings of the IEEE
  Conference on Computer Vision and Pattern Recognition (CVPR)}, 2019.

\bibitem{brock2016generative}
A.~Brock, T.~Lim, J.~M. Ritchie, and N.~Weston, ``Generative and discriminative
  voxel modeling with convolutional neural networks,'' \emph{arXiv preprint
  arXiv:1608.04236}, 2016.

\bibitem{dai2017shape}
A.~Dai, C.~Ruizhongtai~Qi, and M.~Nie{\ss}ner, ``Shape completion using
  3d-encoder-predictor cnns and shape synthesis,'' in \emph{Proceedings of the
  IEEE Conference on Computer Vision and Pattern Recognition}, 2017, pp.
  5868--5877.

\bibitem{litany2018deformable}
O.~Litany, A.~Bronstein, M.~Bronstein, and A.~Makadia, ``Deformable shape
  completion with graph convolutional autoencoders,'' in \emph{Proceedings of
  the IEEE Conference on Computer Vision and Pattern Recognition}, 2018, pp.
  1886--1895.

\bibitem{sinha2017surfnet}
A.~Sinha, A.~Unmesh, Q.~Huang, and K.~Ramani, ``Surfnet: Generating 3d shape
  surfaces using deep residual networks,'' in \emph{Proceedings of the IEEE
  conference on computer vision and pattern recognition}, 2017, pp. 6040--6049.

\bibitem{wu2016learning}
J.~Wu, C.~Zhang, T.~Xue, B.~Freeman, and J.~Tenenbaum, ``Learning a
  probabilistic latent space of object shapes via 3d generative-adversarial
  modeling,'' in \emph{Advances in neural information processing systems},
  2016, pp. 82--90.

\bibitem{smith2017improved}
E.~J. Smith and D.~Meger, ``Improved adversarial systems for 3d object
  generation and reconstruction,'' in \emph{Conference on Robot Learning},
  2017, pp. 87--96.

\bibitem{wang2018pixel2mesh}
N.~Wang, Y.~Zhang, Z.~Li, Y.~Fu, W.~Liu, and Y.-G. Jiang, ``Pixel2mesh:
  Generating 3d mesh models from single rgb images,'' in \emph{Proceedings of
  the European Conference on Computer Vision (ECCV)}, 2018, pp. 52--67.

\bibitem{groueix2018papier}
T.~Groueix, M.~Fisher, V.~G. Kim, B.~C. Russell, and M.~Aubry, ``A
  papier-m{\^a}ch{\'e} approach to learning 3d surface generation,'' in
  \emph{Proceedings of the IEEE conference on computer vision and pattern
  recognition}, 2018, pp. 216--224.

\bibitem{wang20193dn}
W.~Wang, D.~Ceylan, R.~Mech, and U.~Neumann, ``3dn: 3d deformation network,''
  in \emph{Proceedings of the IEEE Conference on Computer Vision and Pattern
  Recognition}, 2019, pp. 1038--1046.

\bibitem{mescheder2019occupancy}
L.~Mescheder, M.~Oechsle, M.~Niemeyer, S.~Nowozin, and A.~Geiger, ``Occupancy
  networks: Learning 3d reconstruction in function space,'' in
  \emph{Proceedings of the IEEE Conference on Computer Vision and Pattern
  Recognition}, 2019, pp. 4460--4470.

\bibitem{park2019deepsdf}
J.~J. Park, P.~Florence, J.~Straub, R.~Newcombe, and S.~Lovegrove, ``Deepsdf:
  Learning continuous signed distance functions for shape representation,'' in
  \emph{Proceedings of the IEEE Conference on Computer Vision and Pattern
  Recognition}, 2019, pp. 165--174.

\bibitem{michalkiewicz2019deep}
M.~Michalkiewicz, J.~K. Pontes, D.~Jack, M.~Baktashmotlagh, and A.~Eriksson,
  ``Deep level sets: Implicit surface representations for 3d shape inference,''
  \emph{arXiv preprint arXiv:1901.06802}, 2019.

\bibitem{cubes1987high}
M.~Cubes, ``A high resolution 3d surface construction algorithm,'' in
  \emph{Proceedings of the 14th Annual Conference on Computer Graphics and
  Interactive Techniques. New York: Association for Computing Machinery}, 1987,
  pp. 163--69.

\bibitem{guo2020deep}
Y.~Guo, H.~Wang, Q.~Hu, H.~Liu, L.~Liu, and M.~Bennamoun, ``Deep learning for
  3d point clouds: A survey,'' \emph{IEEE Transactions on Pattern Analysis and
  Machine Intelligence}, 2020.

\bibitem{yang2019modeling}
J.~Yang, Q.~Zhang, B.~Ni, L.~Li, J.~Liu, M.~Zhou, and Q.~Tian, ``Modeling point
  clouds with self-attention and gumbel subset sampling,'' in \emph{Proceedings
  of the IEEE Conference on Computer Vision and Pattern Recognition}, 2019, pp.
  3323--3332.

\bibitem{zhao2019pointweb}
H.~Zhao, L.~Jiang, C.-W. Fu, and J.~Jia, ``Pointweb: Enhancing local
  neighborhood features for point cloud processing,'' in \emph{Proceedings of
  the IEEE Conference on Computer Vision and Pattern Recognition}, 2019, pp.
  5565--5573.

\bibitem{duan2019structural}
Y.~Duan, Y.~Zheng, J.~Lu, J.~Zhou, and Q.~Tian, ``Structural relational
  reasoning of point clouds,'' in \emph{Proceedings of the IEEE Conference on
  Computer Vision and Pattern Recognition}, 2019, pp. 949--958.

\bibitem{yan2020pointasnl}
X.~Yan, C.~Zheng, Z.~Li, S.~Wang, and S.~Cui, ``Pointasnl: Robust point clouds
  processing using nonlocal neural networks with adaptive sampling,'' in
  \emph{Proceedings of the IEEE/CVF Conference on Computer Vision and Pattern
  Recognition}, 2020, pp. 5589--5598.

\bibitem{li2018pointcnn}
Y.~Li, R.~Bu, M.~Sun, W.~Wu, X.~Di, and B.~Chen, ``Pointcnn: Convolution on
  x-transformed points,'' in \emph{Advances in neural information processing
  systems}, 2018, pp. 820--830.

\bibitem{thomas2019kpconv}
H.~Thomas, C.~R. Qi, J.-E. Deschaud, B.~Marcotegui, F.~Goulette, and L.~J.
  Guibas, ``Kpconv: Flexible and deformable convolution for point clouds,'' in
  \emph{Proceedings of the IEEE International Conference on Computer Vision},
  2019, pp. 6411--6420.

\bibitem{wu2019pointconv}
W.~Wu, Z.~Qi, and L.~Fuxin, ``Pointconv: Deep convolutional networks on 3d
  point clouds,'' in \emph{Proceedings of the IEEE Conference on Computer
  Vision and Pattern Recognition}, 2019, pp. 9621--9630.

\bibitem{shen2018mining}
Y.~Shen, C.~Feng, Y.~Yang, and D.~Tian, ``Mining point cloud local structures
  by kernel correlation and graph pooling,'' in \emph{Proceedings of the IEEE
  conference on computer vision and pattern recognition}, 2018, pp. 4548--4557.

\bibitem{han2017high}
X.~Han, Z.~Li, H.~Huang, E.~Kalogerakis, and Y.~Yu, ``High-resolution shape
  completion using deep neural networks for global structure and local geometry
  inference,'' in \emph{Proceedings of the IEEE International Conference on
  Computer Vision}, 2017, pp. 85--93.

\bibitem{stutz2018learning}
D.~Stutz and A.~Geiger, ``Learning 3d shape completion from laser scan data
  with weak supervision,'' in \emph{Proceedings of the IEEE Conference on
  Computer Vision and Pattern Recognition}, 2018, pp. 1955--1964.

\bibitem{le2018pointgrid}
T.~Le and Y.~Duan, ``Pointgrid: A deep network for 3d shape understanding,'' in
  \emph{Proceedings of the IEEE conference on computer vision and pattern
  recognition}, 2018, pp. 9204--9214.

\bibitem{wang2017cnn}
P.-S. Wang, Y.~Liu, Y.-X. Guo, C.-Y. Sun, and X.~Tong, ``O-cnn: Octree-based
  convolutional neural networks for 3d shape analysis,'' \emph{ACM Transactions
  on Graphics (TOG)}, vol.~36, no.~4, p.~72, 2017.

\bibitem{wang2018adaptive}
P.-S. Wang, C.-Y. Sun, Y.~Liu, and X.~Tong, ``Adaptive o-cnn: a patch-based
  deep representation of 3d shapes,'' in \emph{SIGGRAPH Asia 2018 Technical
  Papers}.\hskip 1em plus 0.5em minus 0.4em\relax ACM, 2018, p. 217.

\bibitem{vakalopoulou2018atlasnet}
M.~Vakalopoulou, G.~Chassagnon, N.~Bus, R.~Marini, E.~I. Zacharaki, M.-P.
  Revel, and N.~Paragios, ``Atlasnet: multi-atlas non-linear deep networks for
  medical image segmentation,'' in \emph{International Conference on Medical
  Image Computing and Computer-Assisted Intervention}.\hskip 1em plus 0.5em
  minus 0.4em\relax Springer, 2018, pp. 658--666.

\bibitem{yang2018foldingnet}
Y.~Yang, C.~Feng, Y.~Shen, and D.~Tian, ``Foldingnet: Point cloud auto-encoder
  via deep grid deformation,'' in \emph{Proceedings of the IEEE Conference on
  Computer Vision and Pattern Recognition}, 2018, pp. 206--215.

\bibitem{liu2020morphing}
M.~Liu, L.~Sheng, S.~Yang, J.~Shao, and S.-M. Hu, ``Morphing and sampling
  network for dense point cloud completion,'' in \emph{Proceedings of the AAAI
  conference on artificial intelligence}, vol.~34, no.~07, 2020, pp.
  11\,596--11\,603.

\bibitem{huang2020pf}
Z.~Huang, Y.~Yu, J.~Xu, F.~Ni, and X.~Le, ``Pf-net: Point fractal network for
  3d point cloud completion,'' in \emph{Proceedings of the IEEE/CVF Conference
  on Computer Vision and Pattern Recognition}, 2020, pp. 7662--7670.

\bibitem{zhang2020detail}
W.~Zhang, Q.~Yan, and C.~Xiao, ``Detail preserved point cloud completion via
  separated feature aggregation,'' in \emph{Proceedings of the European
  Conference on Computer Vision (ECCV)}, 2020, pp. 512--528.

\bibitem{wen2020point}
X.~Wen, T.~Li, Z.~Han, and Y.-S. Liu, ``Point cloud completion by
  skip-attention network with hierarchical folding,'' in \emph{Proceedings of
  the IEEE/CVF Conference on Computer Vision and Pattern Recognition}, 2020,
  pp. 1939--1948.

\bibitem{xie2020grnet}
H.~Xie, H.~Yao, S.~Zhou, J.~Mao, S.~Zhang, and W.~Sun, ``Grnet: Gridding
  residual network for dense point cloud completion,'' in \emph{Proceedings of
  the European Conference on Computer Vision (ECCV)}, 2020.

\bibitem{dai2020sg}
A.~Dai, C.~Diller, and M.~Nie{\ss}ner, ``Sg-nn: Sparse generative neural
  networks for self-supervised scene completion of rgb-d scans,'' in
  \emph{Proceedings of the IEEE/CVF Conference on Computer Vision and Pattern
  Recognition}, 2020, pp. 849--858.

\bibitem{shu20193d}
D.~W. Shu, S.~W. Park, and J.~Kwon, ``3d point cloud generative adversarial
  network based on tree structured graph convolutions,'' in \emph{Proceedings
  of the IEEE/CVF International Conference on Computer Vision}, 2019, pp.
  3859--3868.

\bibitem{zhang2018mixup}
\BIBentryALTinterwordspacing
H.~Zhang, M.~Cisse, Y.~N. Dauphin, and D.~Lopez-Paz, ``mixup: Beyond empirical
  risk minimization,'' in \emph{International Conference on Learning
  Representations}, 2018. [Online]. Available:
  \url{https://openreview.net/forum?id=r1Ddp1-Rb}
\BIBentrySTDinterwordspacing

\bibitem{fan2017point}
H.~Fan, H.~Su, and L.~J. Guibas, ``A point set generation network for 3d object
  reconstruction from a single image,'' in \emph{Proceedings of the IEEE
  conference on computer vision and pattern recognition}, 2017, pp. 605--613.

\bibitem{mao2017least}
X.~Mao, Q.~Li, H.~Xie, R.~Y. Lau, Z.~Wang, and S.~Paul~Smolley, ``Least squares
  generative adversarial networks,'' in \emph{Proceedings of the IEEE
  International Conference on Computer Vision}, 2017, pp. 2794--2802.

\bibitem{wen2021pmp}
X.~Wen, P.~Xiang, Z.~Han, Y.-P. Cao, P.~Wan, W.~Zheng, and Y.-S. Liu,
  ``Pmp-net: Point cloud completion by learning multi-step point moving
  paths,'' in \emph{Proceedings of the IEEE/CVF Conference on Computer Vision
  and Pattern Recognition}, 2021, pp. 7443--7452.

\bibitem{wang2020softpoolnet}
Y.~Wang, D.~J. Tan, N.~Navab, and F.~Tombari, ``Softpoolnet: Shape descriptor
  for point cloud completion and classification,'' \emph{Proceedings of the
  European Conference on Computer Vision (ECCV)}, 2020.

\bibitem{Chen2020Unpaired}
\BIBentryALTinterwordspacing
X.~Chen, B.~Chen, and N.~J. Mitra, ``Unpaired point cloud completion on real
  scans using adversarial training,'' in \emph{International Conference on
  Learning Representations}, 2020. [Online]. Available:
  \url{https://openreview.net/forum?id=HkgrZ0EYwB}
\BIBentrySTDinterwordspacing

\bibitem{tatarchenko2019single}
M.~Tatarchenko, S.~R. Richter, R.~Ranftl, Z.~Li, V.~Koltun, and T.~Brox, ``What
  do single-view 3d reconstruction networks learn?'' in \emph{Proceedings of
  the IEEE Conference on Computer Vision and Pattern Recognition}, 2019, pp.
  3405--3414.

\bibitem{wang2020cascaded}
X.~Wang, M.~H. Ang~Jr, and G.~H. Lee, ``Cascaded refinement network for point
  cloud completion,'' in \emph{Proceedings of the IEEE/CVF Conference on
  Computer Vision and Pattern Recognition}, 2020, pp. 790--799.

\bibitem{sarmad2019rl}
M.~Sarmad, H.~J. Lee, and Y.~M. Kim, ``Rl-gan-net: A reinforcement learning
  agent controlled gan network for real-time point cloud shape completion,'' in
  \emph{Proceedings of the IEEE Conference on Computer Vision and Pattern
  Recognition}, 2019, pp. 5898--5907.

\bibitem{richard2020kaplan}
A.~Richard, I.~Cherabier, M.~R. Oswald, M.~Pollefeys, and K.~Schindler,
  ``Kaplan: A 3d point descriptor for shape completion,'' in \emph{2020
  International Conference on 3D Vision (3DV)}.\hskip 1em plus 0.5em minus
  0.4em\relax IEEE, 2020, pp. 101--110.

\bibitem{wu20153d}
Z.~Wu, S.~Song, A.~Khosla, F.~Yu, L.~Zhang, X.~Tang, and J.~Xiao, ``3d
  shapenets: A deep representation for volumetric shapes,'' in
  \emph{Proceedings of the IEEE conference on computer vision and pattern
  recognition}, 2015, pp. 1912--1920.

\bibitem{geiger2013vision}
A.~Geiger, P.~Lenz, C.~Stiller, and R.~Urtasun, ``Vision meets robotics: The
  kitti dataset,'' \emph{The International Journal of Robotics Research},
  vol.~32, no.~11, pp. 1231--1237, 2013.

\bibitem{Matterport3D}
A.~Chang, A.~Dai, T.~Funkhouser, M.~Halber, M.~Niessner, M.~Savva, S.~Song,
  A.~Zeng, and Y.~Zhang, ``{Matterport3D}: Learning from {RGB-D} data in indoor
  environments,'' \emph{International Conference on 3D Vision (3DV)}, 2017.

\bibitem{zhang2021point}
J.~Zhang, W.~Chen, Y.~Wang, R.~Vasudevan, and M.~Johnson-Roberson, ``Point set
  voting for partial point cloud analysis,'' \emph{IEEE Robotics and Automation
  Letters}, vol.~6, no.~2, pp. 596--603, 2021.

\bibitem{groueix2018}
T.~Groueix, M.~Fisher, V.~G. Kim, B.~Russell, and M.~Aubry, ``{AtlasNet: A
  Papier-M\^ach\'e Approach to Learning 3D Surface Generation},'' in
  \emph{Proceedings IEEE Conf. on Computer Vision and Pattern Recognition
  (CVPR)}, 2018.

\bibitem{kingma2014adam}
D.~P. Kingma and J.~Ba, ``Adam: A method for stochastic optimization,''
  \emph{arXiv preprint arXiv:1412.6980}, 2014.

\bibitem{heusel2017gans}
M.~Heusel, H.~Ramsauer, T.~Unterthiner, B.~Nessler, and S.~Hochreiter, ``Gans
  trained by a two time-scale update rule converge to a local nash
  equilibrium,'' in \emph{Advances in Neural Information Processing Systems},
  2017, pp. 6626--6637.

\bibitem{kanazawa2018end}
A.~Kanazawa, M.~J. Black, D.~W. Jacobs, and J.~Malik, ``End-to-end recovery of
  human shape and pose,'' in \emph{Proceedings of the IEEE Conference on
  Computer Vision and Pattern Recognition}, 2018, pp. 7122--7131.

\bibitem{qin2019pointdan}
C.~Qin, H.~You, L.~Wang, C.-C.~J. Kuo, and Y.~Fu, ``Pointdan: A multi-scale 3d
  domain adaption network for point cloud representation,'' in \emph{Advances
  in Neural Information Processing Systems}, 2019, pp. 7192--7203.

\end{thebibliography}


%

\vspace{-14mm}
\begin{IEEEbiography}[{\includegraphics[width=1in,height=1.25in,clip,keepaspectratio]{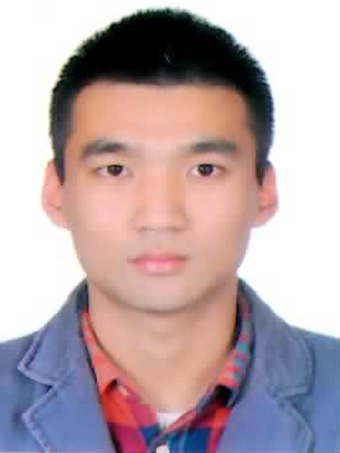}}]{Xiaogang Wang} is currently a Research Fellow in National University of Singapore, where he obtained his Ph.D degree in Mechanical Engineering in 2021. He obtained his M.Eng. degree in Mechanical Engineering from the Beihang University in 2016; and the B.Sc. degree in Mechanical Engineering from the University of Science and Technology Beijing in 2013, respectively. His research interests are computer vision and deep learning.
\end{IEEEbiography}

\vspace{-16mm}
\begin{IEEEbiography}[{\includegraphics[width=1in,height=1.25in,clip,keepaspectratio]{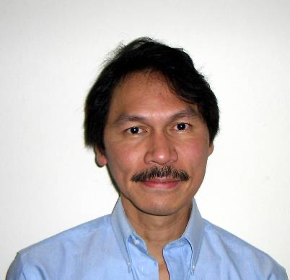}}] {Marcelo H Ang Jr} is a Professor in the Department of Mechanical Engineering, National University of Singapore. He received the B.Sc. degrees (Cum Laude) in Mechanical Engineering and Industrial Management Engineering from the De La Salle University, Manila, Philippines, in 1981; the M.Sc. degree in Mechanical Engineering from the University of Hawaii at Manoa, Honolulu, Hawaii, in 1985; and the M.Sc. and Ph.D. degrees in Electrical Engineering from the University of Rochester, Rochester, New York, in 1986 and 1988, respectively. His work experience includes heading the Technical Training Division of Intel's Assembly and Test Facility in the Philippines, research positions at the East West Center in Hawaii and at the Massachusetts Institute of Technology, and a faculty position as an Assistant Professor of Electrical Engineering at the University of Rochester, New York. In 1989, Dr. Ang joined the Department of Mechanical Engineering of the National University of Singapore, where he is currently a Professor. He is also the Acting Director of the Advanced Robotics Centre. His research interests span the areas of robotics, mechatronics, and applications of intelligent systems methodologies. He teaches both at the graduate and undergraduate levels in the following areas: robotics; creativity and innovation, and Engineering Mathematics. He is also active in consulting work in robotics and intelligent systems. In addition to academic and research activities, he is actively involved in the Singapore Robotic Games as its founding chairman and the World Robot Olympiad as a member of the Advisory Council.
\end{IEEEbiography}

\vspace{-14mm}
\begin{IEEEbiography}[{\includegraphics[width=1in,height=1.25in,clip,keepaspectratio]{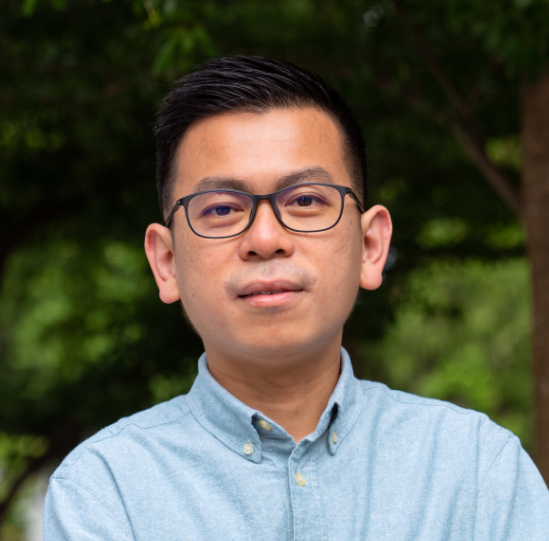}}]{Gim Hee Lee} is currently an Assistant Professor at the Department of Computer Science at the National University of Singapore (NUS), where he heads the Computer Vision and Robotic Perception (CVRP) Laboratory. He is also affiliated with the NUS Graduate School for Integrative Sciences and Engineering (NGS), and the NUS Institute of Data Science (IDS). He was a researcher at Mitsubishi Electric Research Laboratories (MERL), USA. Prior to MERL, he did his PhD in Computer Science at ETH Zurich. He received his B.Eng with first class honors and M.Eng degrees from the Department of Mechanical Engineering at NUS. His research interests include computer vision, robotic perception and machine learning. He has served as an Area Chair for BMVC 2020, 3DV 2020, CVPR 2021, ICCV 2021, BMVC 2021 and WACV 2022, and will be part of the organizing committee as the Exhibition/Demo Chair for CVPR 2023.
\end{IEEEbiography}




\twocolumn[{%
		\renewcommand\twocolumn[1][]{#1}%
		\vskip .5in
		\begin{center}
			\textbf{\Large Cascaded Refinement Network for Point Cloud Completion with Self-supervision - }\\
			\vspace*{6pt}
			\textbf{\Large  Supplementary Material}\\
			\vspace*{10pt}
			{\large
				Xiaogang Wang,
        Marcelo H Ang Jr,~\IEEEmembership{Senior Member,~IEEE,} 
        and~Gim Hee Lee,~\IEEEmembership{Member,~IEEE}\\
			}
			\vskip .5em
			\vspace*{10pt}
		\end{center}
	}]

\setcounter{equation}{0}
\setcounter{figure}{0}
\setcounter{table}{0}
\setcounter{section}{0}

\maketitle

\IEEEdisplaynontitleabstractindextext

\ifCLASSOPTIONpeerreview
\begin{center} \bfseries EDICS Category: 3-BBND \end{center}
\fi
%
\IEEEpeerreviewmaketitle

\section{Network Architecture Details}

We illustrate the details of the coarse reconstruction, dense reconstruction and discriminator in this section.

\subsection{Coarse Reconstruction}
The coarse reconstruction stage consists of three fully-connected layers with the size of 1024, 1024 and 1536 units and a reshaping operation to obtain $512\times 3$ points.

\subsection{Dense Reconstruction}

We design a lifting module to increase the point size by a factor of two and concurrently refine the point positions by the feature contraction-expansion unit.
Shared multi-layer perceptrons (MLPs) are used for feature contraction and expansion due to its efficiency.
The two operations are represented as $f_c=\mathcal{RS}(\mathcal C_{C}(f_S))$ and $f_e=\mathcal{RS}(\mathcal C_{E}(f_c))$, respectively. $\mathcal{RS}(.)$ is a reshaping operation. 
$\mathcal C_{C}(\cdot)$ and $\mathcal C_{E}(\cdot)$ are two MLPs blocks with $\{64,128\}$ and $\{64\}$ neurons, respectively.
$f_c$ is the feature obtained by contraction, and $f_e$ is the feature obtained by expansion.
Our lifting module predicts point feature residuals instead of the final output since deep neural networks are better at predicting residuals. 
Overall, the output point set $\text{P}_{i}$ in an one-step refinement is represented as:
\begin{align}
    \text{P}_i&=F(\text{P}_S^{\prime})+\text{P}_S^{\prime},
\end{align}
where $F(.)$ represents the lifting module, and it predicts per-vertex offsets for the input point set $\text{P}_S^{\prime}$.
ReLU activation is used for all the convolution layers except for the last one.

\subsection{Discriminator}
\begin{figure}
\centering
  \includegraphics[width=1\linewidth]{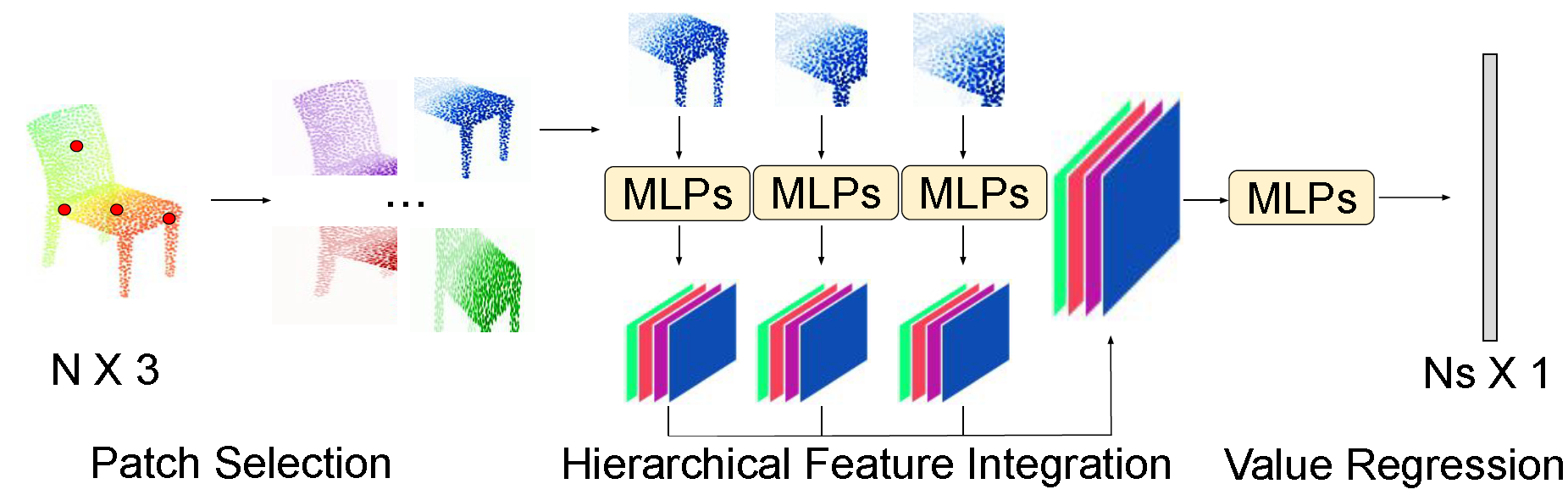}
  \caption{The discriminator architecture sub-network. It includes the patch selection, hierarchical feature integration and confidence value regression.}
  \label{discriminator}
\end{figure}

As shown in Figure~\ref{discriminator}, our discriminator consists of a patch selection, a hierarchical feature integration and a value regression. 

\myparagraph{Patch selection.} We first sample 256 seed points by FPS from the complete point sets similar with PointNet++. 
Neighboring points that are within a radius to the seed points are collected by the ball query method.
The three radii are $\{0.1, 0.2, 0.4\}$ and the number of neighborhoods $K$s are $\{16, 32, 128\}$.

\myparagraph{Hierarchical feature integration.} The inputs are 256 local regions of points with data size of $ 256 \times K \times 3$. Three shared MLPs blocks with $\{16, 16, 32\}$, $\{32, 32, 64\}$ and $\{32, 48, 64\}$ neurons followed by a max-pooling operation are used to extract locally global point features. The output is the concatenation of three $ 256 \times 160$ features.

\myparagraph{Value regression.} The confidence scores for each local patch are regressed by an MLP with one single neuron. The output size is $ 256 \times 1$.

\section{More Experimental Results}
\subsection{Qualitative Results of non-symmetric Objects}
\begin{figure}
\centering
  \includegraphics[width=1\linewidth]{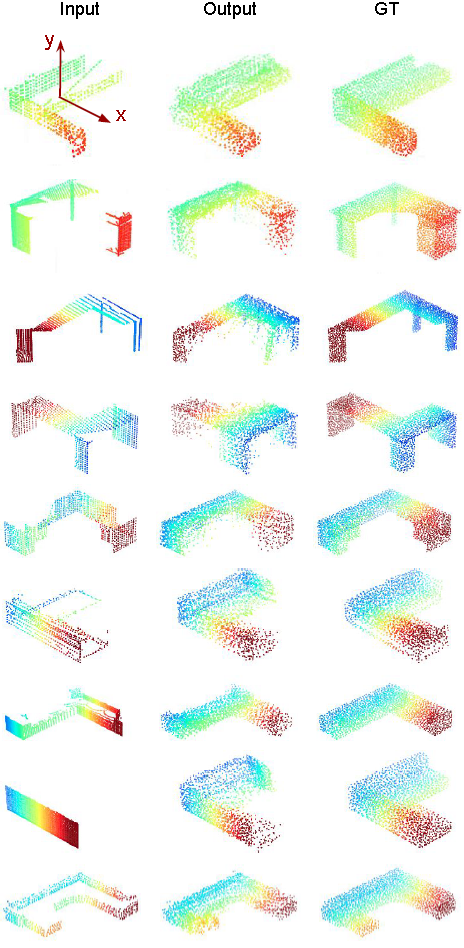}
  \caption{Qualitative results of non-symmetric objects.}
  \label{non_symmetric_objects}
\end{figure}

Qualitative results of the non-symmetric objects are shown in Figure~\ref{non_symmetric_objects}, where reasonable object shapes with high-quality details are generated. 
This verifies that although we assume the objects are symmetrical regarding \emph{xy}-plane, our network is not limited to the symmetrical objects. 
We mirror the inputs to provide possible initial point coordinates for the lifting module. The mirrored points may not belong to the missing part, but the coordinates are further refined by our lifting module.

\subsection{More Qualitative Results}
\begin{figure*}
\centering
  \includegraphics[width=0.95\linewidth]{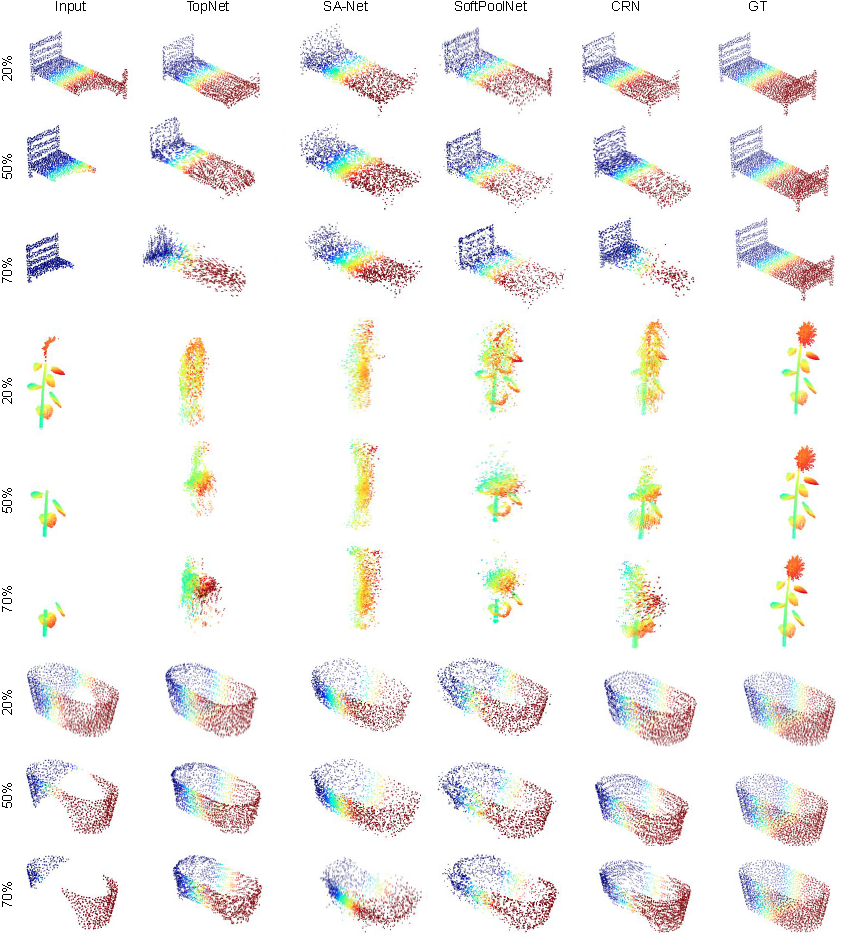}
  \caption{Qualitative comparison results for the occluded ModelNet dataset. We show different results for three occlusion ratios: $20\%$, $50\%$ and $70\%$. The resolutions of output and ground truth are 2048.}
  \label{occlusion_modelnet}
\end{figure*}

Figures~\ref{occlusion_modelnet} shows the qualitative results for the occlusion experiments. We remove different ratios of points from the complete point sets to test the robustness of the methods. Specifically, we remove 20$\%$, 50$\%$ and 70$\%$ of points from the complete point clouds. The performances of different methods drop with the increase in occlusion ratios. Nonetheless, our results are more realistic compared to other approaches.

\end{document}